\documentclass{article}

\usepackage{microtype}
\usepackage{graphicx}
\usepackage{subcaption}
\usepackage{booktabs} % for professional tables

\usepackage{hyperref}

\usepackage[preprint]{icml2026}

\usepackage{amsmath}
\usepackage{amssymb}
\usepackage{mathtools}
\usepackage{amsthm}

\usepackage[utf8]{inputenc} % allow utf-8 input
\usepackage[T1]{fontenc}    % use 8-bit T1 fonts
\usepackage{hyperref}       % hyperlinks
\usepackage{url}            % simple URL typesetting
\usepackage{booktabs}       % professional-quality tables
\usepackage{amsfonts}       % blackboard math symbols
\usepackage{nicefrac}       % compact symbols for 1/2, etc.
\usepackage{microtype}      % microtypography
\usepackage{xcolor}         % colors
\usepackage[table]{xcolor}
\usepackage{multirow}
\usepackage{booktabs}
\usepackage{amsmath}
\usepackage{todonotes}
\usepackage{algorithmic,algorithm}
\usepackage{framed}
\usepackage{wrapfig}
\usepackage{enumitem}

\definecolor{babyblue}{HTML}{4285f4}
\definecolor{LightGrey}{HTML}{d9d9d9}

\newcommand{\method}{{DP-RFT}} 
\newcommand{\augpe}{{Aug-PE}}

\usepackage[capitalize,noabbrev]{cleveref}

\theoremstyle{plain}

\theoremstyle{definition}

\theoremstyle{remark}

\icmltitlerunning{DP-RFT: Learning to Generate Synthetic Text via Differentially Private Reinforcement Fine-Tuning}

\begin{document}

\twocolumn[
  \icmltitle{DP-RFT: Learning to Generate Synthetic Text \\
  via Differentially Private Reinforcement Fine-Tuning}



  % \icmlsetsymbol{intern}{*}

  % Define a custom symbol

  \begin{icmlauthorlist}
    \icmlauthor{Fangyuan Xu*}{nyu}
    \icmlauthor{Sihao Chen}{msft}
    \icmlauthor{Zinan Lin}{msft}
    \icmlauthor{Taiwei Shi}{usc}
    \icmlauthor{Sydney Graham}{msft}
    \icmlauthor{Pei Zhou}{msft}
    \icmlauthor{Mengting Wan}{msft}
    \icmlauthor{Alex Stein}{msft}
    \icmlauthor{Virginia Estellers}{msft}
    \icmlauthor{Charles Chen}{msft}
    \icmlauthor{Morris Sharp}{msft}
    \icmlauthor{Richard Speyer}{msft}
    \icmlauthor{Tadas Baltrusaitis}{msft}
    \icmlauthor{Jennifer Neville}{msft}
    \icmlauthor{Eunsol Choi}{nyu}
    \icmlauthor{Longqi Yang}{msft}
  \end{icmlauthorlist}

  \icmlaffiliation{nyu}{New York University}
  \icmlaffiliation{msft}{Microsoft}
  \icmlaffiliation{usc}{University of Southern California}

  \icmlcorrespondingauthor{Fangyuan Xu}{fx2079@nyu.edu}
  \icmlcorrespondingauthor{Sihao Chen}{sihaochen@microsoft.com}

  % \renewcommand{\thefootnote}{*}
  % \footnotetext{Work done during an internship at Microsoft.}
  \icmlkeywords{Machine Learning, ICML}

  \vskip 0.3in
]



\printAffiliationsAndNotice{*Work done during an internship at Microsoft.}  % no special notice (required even if empty)

\begin{abstract}
Differentially private (DP) synthetic data generation plays a pivotal role in developing large language models (LLMs) on private data, where data owners cannot provide eyes-on access to individual examples. 
Generating DP synthetic data typically involves a difficult trade-off. On one hand, DP finetuning methods train an LLM as a synthetic data generator with formal privacy guarantees, yet it still requires the raw content of private examples for model training. However, methods that avoid direct exposure to private data are bounded by an off-the-shelf, un-finetuned model, whose outputs often lack domain fidelity. 
\textbf{Can we \textit{train} an LLM to generate high-quality synthetic text without eyes-on access to individual private examples? } 
In this work, we introduce \textit{Differentially Private Reinforcement Fine-Tuning} (DP-RFT), an online reinforcement learning algorithm for synthetic data generation with LLMs. DP-RFT leverages DP-protected nearest-neighbor votes from an eyes-off private corpus as a reward signal for on-policy synthetic samples generated by an LLM. The LLM iteratively learns to generate synthetic data to maximize the expected DP votes through Proximal Policy Optimization (PPO). We evaluate DP-RFT for long-form and domain-specific synthetic data generation, such as news articles, meeting transcripts, and medical article abstracts. 
Our experiments show that DP-RFT closes the gap between private evolution and DP finetuning methods in terms of the fidelity and downstream utility of the generated synthetic data, while respecting the private data boundary.
\end{abstract}

\section{Introduction}
% Large language models training today happens in centralized setting, where data providers 

Large language models (LLMs) have achieved remarkable progress in their capabilities through massive-scale training \citep{kaplan2020scaling, biderman2023pythia}.
However, LLM training requires centralized access to its training data. This is typically incompatible with \textit{private data}, where data owners cannot provide eyes-on access to the data due to privacy constraints, regulations, or other reasons \cite{Carlini2020ExtractingTD, carlini2024stealing, pandaprivacy, Chen2025ASO}. 

% However, much of the non-public data not yet seen in pre-training lies is data stores that are currently inaccessible to model training, due at least in part to privacy concerns from the data owners (CITE CITE) and regulators.
% However, as LLMs are increasingly trained on real-world data, there are growing concerns around the responsible use of customer or copyrighted data \citep{carlini2024stealing, pandaprivacy}. 
% LLMs trained on private or sensitive data can memorize these patterns, causing security and privacy risks by exposing private information through the model outputs \citep{biderman2023emergent, chen2024copybench, shi2024detecting}.

\textit{Differentially-private} (DP) synthetic data has emerged as a responsible way of circumventing the need for eyes-on data access. The goal is to leverage LLMs to generate a synthetic dataset that is statistically similar to the private dataset \citep{jordon2018pate, li2022large, yu2022differentially, yue-etal-2023-synthetic, harder2023pre}, while ensuring the privacy guarantee that individual samples in the original data cannot be inferred from the model outputs \citep{dwork2006calibrating}. DP synthetic data can be used for downstream model development 
% (e.g. training, evaluation, etc.) 
with intact privacy guarantee thanks to DP's post-processing properties \citep{dwork2014algorithmic}. 

Existing DP synthetic data generation methods typically face a difficult trade-off between data quality versus the required level of access to private dataset during data generation. 
While DP finetuning methods, e.g. DP-SGD~\cite{Abadi2016DeepLW}, provide privacy guarantee when training the synthetic data generator on the private dataset, they still require data custodians to share and expose raw private data samples in the training process \citep{pe1, ponomareva2023dp}.  
% direct access to raw private data during training. This violates the \textit{eyes-off} constraints where data custodians cannot expose or share individual records . 
On the other hand, 
% to accommodate such common constraints in enterprise setting~\citep{apple2025differentialprivacy}, 
methods which avoid direct exposure to private data e.g. \augpe{} \citep{xie2024differentially}, leverage iterative prompting to guide off-the-shelf LLMs to generate data that are more similar to the private distribution. However, as such approaches are bounded by samples generated from a frozen LLM, the outputs often lack domain fidelity.

% Using LLM to generate DP synthetic text often involves a difficult trade-off. On one hand, existing methods that rely on training or finetuning against private dataset, e.g. DP-SGD \citep{Abadi2016DeepLW}, usually require direct access to raw private data during training, which may be infeasible under ``eyes-off'' constraints settings where data custodians cannot expose or share individual records \citep{pe1, ponomareva2023dp, apple2025differentialprivacy}. On the other hand, existing methods that avoid direct exposure to private data, e.g. AugPE \citep{xie2024differentially}, are bounded by samples generated from an off-the-shelf, frozen LLM, whose outputs often lack domain fidelity.

To get the best of both worlds, we propose \emph{Differentially Private Reinforcement Fine-Tuning} (DP-RFT), a reinforcement learning (RL) algorithm for training LLMs as DP synthetic data generators without the need to expose individual private examples to LLM during the process. Building upon the reward-based RL post-training techniques for LLMs \citep{Ouyang2022TrainingLM, zelikman2022star, jaech2024openai, guo2025deepseek}, DP-RFT leverages DP-protected nearest-neighbor votes (DP Votes, \citet{pe1, xie2024differentially}) from an eyes-off private corpus as a reward signal for LLM's on-policy synthetic sample generation. The DP-votes can be calculated within the boundary of the private data and then exposed to LLM with formal privacy guarantee, so that the private data owners do not need to provide access for individual private examples to LLM during training. The LLM learns from reward feedback to generate synthetic data to maximize the expected DP votes through Proximal Policy Optimization (PPO; \citeauthor{Schulman2017ProximalPO}, \citeyear{Schulman2017ProximalPO}). 

We evaluate DP-RFT for \textit{long-form} synthetic data generation with different domains and types of text, such as news articles~\citep{Narayan2018DontGM}, meeting transcripts \citep{zhong-etal-2021-qmsum}, LLM-user chat logs~\citep{zhao2024wildchat} and medical article abstracts \citep{yu2023training}. We train DP-RFT models under different levels of privacy budget, and evaluate the generated synthetic data by intrinsic quality (\S~\ref{ssec:sim_eval}), as well as downstream utility when the synthetic data is used as training data for another language model (\S~\ref{ssec:down_eval}).
Our experiments show that DP-RFT outperforms Aug-PE in terms of the fidelity and downstream utility of the generated synthetic data, and closes the gap between DP finetuning and Aug-PE in terms of downstream utility. Our key contributions are: 
\begin{enumerate}[leftmargin=*]
    \item \emph{New finetuning method for DP synthetic data generation}:  We propose \method~, which leverages DP-protected reward function and  reinforcement learning to train a LLM against private corpus, without \textit{ever} ingesting private examples directly during the training process and hence satisfying the \textit{eyes-off} constraint.
    % a method to fine-tune language models to generate synthetic text that is similar to a private corpus with differential privacy guarantee.%\emph{DP+RL}
    \item \emph{In-depth experiments and analyses}: We conduct comprehensive evaluation of our methods on four datasets, showing significant improvement on downstream utility, especially for a tight privacy budget and on dataset which are out-of-distribution for the backbone language model. Our qualitative analysis further reveals that ~\method~is able to better capture lexical and structural similarity of long-form, structural outputs. %\emph{Experimental results}: 
\end{enumerate}

\begin{figure*}[t!]
    \centering
    \includegraphics[scale=0.175, trim={20 0 0 0},clip]{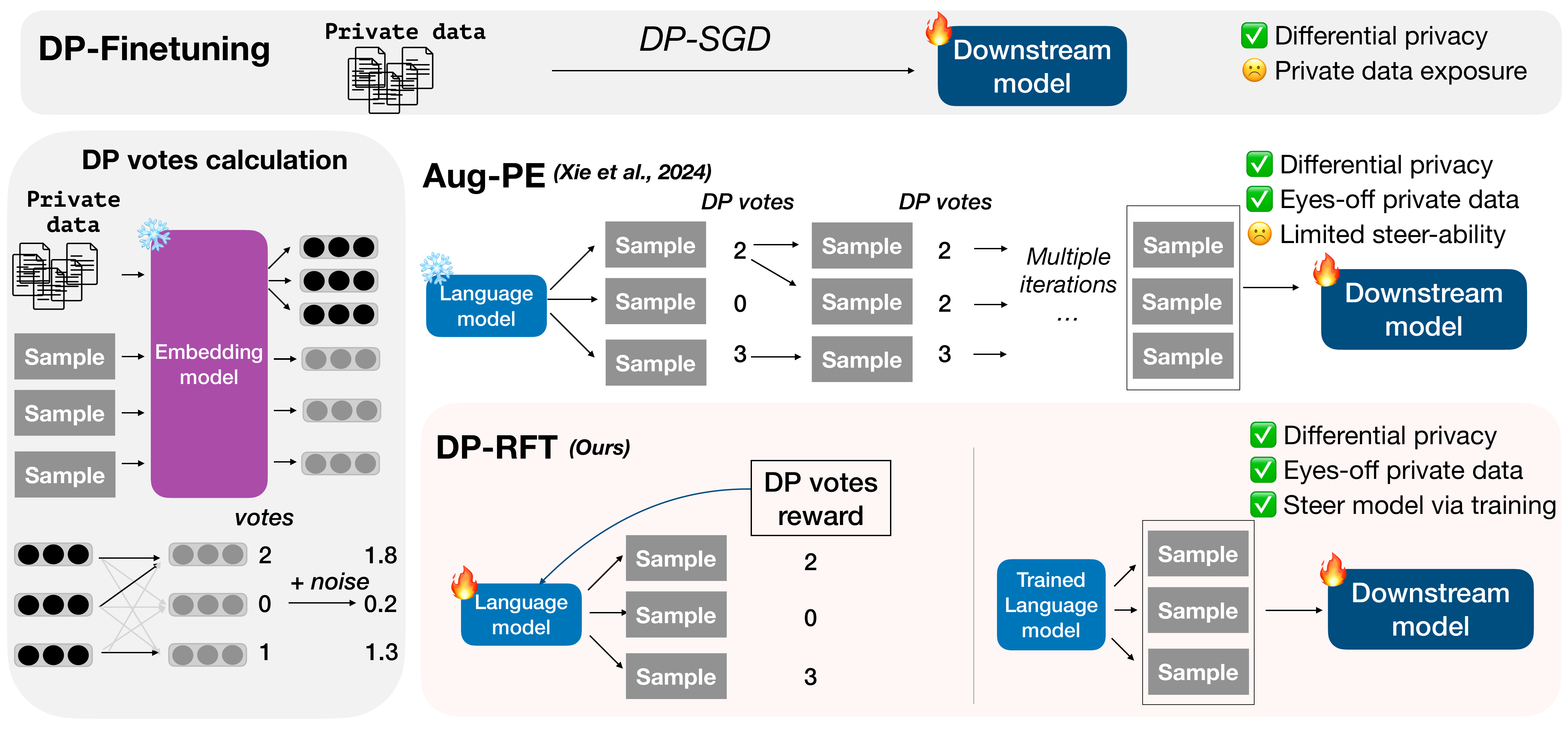}
    \caption{An illustration of \method{} and comparison with prior methods. \method{} fine-tunes a language model (LM) to generate texts similar to the private document with reinforcement learning, guided by a DP-protected nearest-neighbor votes as reward. Unlike DP-Finetuning which requires accessing the private data directly, \method{} and ~\augpe{} allow downstream model training outside of the private boundary. While ~\augpe{} is bounded by samples generated from a frozen LM, ~\method{} generates synthetic data with higher fidelity by training an LM to generate texts more similar to the private data.} 
    \label{fig:intro}
\end{figure*}

\section{Background and Related work}

\paragraph{Differential Privacy with LLMs.} As LLMs are increasingly deployed in many real-world cases, preserving the privacy of training and evaluation data has been an active research area. One popular approach is to apply DP-SGD ~\citep{Abadi2016DeepLW} to train LLMs %either  
% language models 
for classification~\citep{Yu2021DifferentiallyPF, Li2021LargeLM} or %a synthetic data generator model 
generation tasks~\citep{Wang2025RewardDSPF,Yu2021DifferentiallyPF, Li2021LargeLM,yue-etal-2023-synthetic, mattern-etal-2022-differentially, Kurakin2023HarnessingLM, Ngong2024DifferentiallyPL, Tan2025SynthesizingPT}. %However, training large models with DP can be practically challenging, particularly due to the implementation complexity and resource overhead of DP-SGD. 
Later work proposed training-free approaches, such as injecting calibrated DP noise into the token-by-token generation process of LLMs~\citep{tang2023privacy,flemings2025differentially,duan2023flocks}, or aggregating multiple LLM outputs in a DP manner~\citep{wu2023privacy}.
While these methods provide theoretical DP guarantees, they all require exposing the private data as input to the LLM. %either the downstream model or the synthetic data generator. 
In practice, there could be regulatory constraints that require the private data to be completely \textit{eyes-off}, preventing the adoption of such methods. Private Evolution (PE)~\citep{pe1, xie2024differentially, Lin2025DifferentiallyPS,wang2025struct} is a newly emerged framework to address this challenge. 
% Another line of work proposes to prompt large language model to synthetic private data. Private Evolution~\citep{pe1, xie2024differentially, Lin2025DifferentiallyPS} is a line of work which 
It proposes an iterative prompting pipeline which steers the model generation to be more similar to the private data, as measured by embedding similarity. 
While these methods avoid private inputs, their reliance on un-finetuned models limits the synthetic data quality. Our \method{} aims to take the best of both worlds: adapting model weights to better fit private data without directly inputting it into the model. {Recent work~\citep{nguyen2025synthetic} proposes to generate synthetic data via gradient matching, which requires gradient computation on private data.} Notably, \citet{popri2025} recently proposed fine-tuning LLMs using embedding similarity as a reward with an offline reinforcement learning algorithm (DPO; \citeauthor{Rafailov2023DirectPO}, \citeyear{Rafailov2023DirectPO}).

\paragraph{Reinforcement fine-tuning (RFT) for LLMs.} 
We briefly review how reinforcement learning is used to post-train large language models. Let a language model with parameters $\theta$ define a policy $\pi_\theta(\cdot \mid p)$ over token sequences given a prompt $p$. The goal of RFT is to update $\theta$ so that samples $d \sim \pi_\theta(\cdot \mid p)$ have higher task-specific utility, quantified by a scalar reward $R(d,p)$. In modern LLM post-training, $R$ can come from a learned reward model (e.g., RLHF, \citeauthor{Ouyang2022TrainingLM}, \citeyear{Ouyang2022TrainingLM}), or from verifiable signals (e.g., passing tests in code/math), and training proceeds by maximizing the cumulative rewards. Practically, RFT alternates between: (i) sampling on-policy generations from $\pi_\theta$ for a batch of prompts; (ii) computing scalar rewards for each sample; and (iii) updating $\theta$ via a policy gradient method (e.g., PPO, \citeauthor{Schulman2017ProximalPO}, \citeyear{Schulman2017ProximalPO}; or GRPO, \citeauthor{Shao2024DeepSeekMathPT}, \citeyear{Shao2024DeepSeekMathPT}) on those samples. Compared to supervised fine-tuning (SFT), where LLM ingests reference texts as input during training, RFT directly optimizes the task metric exposed by $R$ and is thus effective when high-quality labels are scarce or when a reward function $R$ can be defined. The fact that RFT does not ingest example outputs during training is the key to our DP-RFT algorithm's design.   
% A line of work applies reinforcement learning with a preference model learned from human feedback as reward~\citep{Ouyang2022TrainingLM, Rafailov2023DirectPO}, finding further improvement compared to conducting supervised fine-tuning (SFT) alone. 
% Recently, a line of work propose to directly conducting RFT, bypassing the need to collect SFT data~\citep{DeepSeekAI2025DeepSeekR1IR, zeng2025simplerlzoo, Zhou2025R1ZerosM}. Methods in this space predominately focus on domains with \textit{verifiable} rewards, such as solving mathematical problems~\citep{Shao2024DeepSeekMathPT} and building coding agents~\citep{Wang2024EnhancingCL, Wei2025SWERLAL, zeng-etal-2025-acecoder}. Our work propose an RFT method that can be applied without SFT in a domain \textit{without verifiable rewards}, i.e. synthetic document generation. Recent work has proposed methods for \textit{unverifiable} reward in other domains: \citet{Gurung2025LearningTR} trains a language model to generate plans for a frozen language model which generates a story. They propose to leverage perplexity of the frozen language model on the gold continuation as the reward. ~\citet{Chen2025LearningTR} proposes to conduct RFT to improve factuality of long-form generation, using a combination of factuality score and answer relevance as reward, similar to our setting.

% \paragraph{Synthetic data for LLM.} Using large language models to generate synthetic data is an active research area \citep{wang2023self, Hsieh2023DistillingSO}. \fy{todo}

\section{DP-RFT: Differentially-Private Reinforcement Finetuning}\label{sec:method}

Given a set of private documents $D$\textsubscript{priv}, our goal is to train a language model $M$\textsubscript{gen} that generates synthetic documents $D$\textsubscript{syn} similar to $D$\textsubscript{priv}. $M$\textsubscript{gen} takes an input prompt \textit{p} which contains public information about the private corpus and produces a synthetic document $d$. We describe our method, \method, which leverages reinforcement fine-tuning (RFT) to train $M$\textsubscript{gen} against a DP-protected reward function $R$. We illustrate the workflow of ~\method{}, as well as its differences from prior approaches, including DP-Finetuning \cite{Abadi2016DeepLW} and Augmented Private Evolution (Aug-PE; \citeyear{xie2024differentially}) in \autoref{fig:intro}. 

% \subsection{Reinforcement fine-tuning (RFT) for LLMs.}

\subsection{Generating input prompts}
We use a large language model (LLM) to construct input prompts which contain public information about the target domain. 
We construct a diverse set of prompts to encourage diverse generation from $M$\textsubscript{gen} following Augmented Private Evolution (Aug-PE) \citep{xie2024differentially} for training and inference of \method. We focus on one or both of the two axes of diversity (1) content diversity and (2) length diversity. For content diversity, we generate a set of keywords using an LLM. For instance, for BBC articles, we prompt an LLM to generate a list of topics (e.g. Politics, Sports), along with keywords associated with each topic. We obtain a set of prompts, each with a different set of keywords, which we use as input to $M_{gen}$.  For length diversity, we define a possible range of length for the private documents and random sample one length to include in the prompt (e.g. ``The generated document should contain around 300 words''). We include the details of the procedure in \S \ref{app:implementation_details} in the appendix.

\subsection{Rewards}\label{subsec:reward_design}
\paragraph{DP voting as reward.} Aug-PE \citep{xie2024differentially} prompts language models to generate multiple synthetic documents and iteratively generate synthetic documents via a process called \textbf{DP voting}: each private document votes for a synthetic document that is most similar to them measured with a similarity function, such as a text embedding model. They find that selecting synthetic documents that receive more votes (i.e. close to more private documents in the embedding space) lead to better synthetic documents and hence downstream performance. 
We take inspiration from Aug-PE and propose the following new designs to the DP-voting mechanism and the iterative generation-sampling process:
\textbf{(1) Using raw similarity scores as votes.} In Aug-PE, each private sample votes only for its closest synthetic document, producing a binary (1 or 0) signal. This discards valuable information in the raw similarity scores, such as \textit{how much} one synthetic sample is better than another. To capture this richer signal, we instead use the raw similarity scores $sim(d, D_{priv, i})$ directly as votes from the $i$-th private document $D_{priv, i}$ to the synthetic document $d$, where $sim$ is a similarity measure (e.g. text embedding similarity). 
\textbf{(2) DP-RFT.}
While Aug-PE can promote synthetic documents that are similar to the private documents, the model is untrained, and hence has limited steerability.  We instead leverage the votes as the reward in Reinforcement Fine-Tuning (RFT) to steer $M$\textsubscript{gen} to generate synthetic documents more similar to the private corpus. Concretely, for a synthetic document $d$ and a predefined similarity measure $sim$,  we define $R_{sim} = \frac{1}{{|D_{priv}|}} \sum_{i=1}^{i=|D_{priv}|} sim(d, D_{priv, i})$ as the reward.

%%%%%%%%%%%%%%%% pseudocode for DP-voting as reward %%%%%%%%%%%%%%%%%
\begin{figure}
\begin{framed}
\small
\textbf{Input:} A set of $s$ synthetic documents $D_{syn}$, a similarity metric $sim$, a set of $p$ private documents $D_{priv}$, a noise multiplier $\sigma$, a raw similarity threshold $c$.

\textbf{Output:} Reward $R_{sim}$ for synthetic documents $D_{syn}$

\begin{algorithmic}[1]
\STATE $R_{sim}$$ \gets \left[ \, \right]$
\FOR{$i  \in \{1, \dots, s \}$} % \zinan{I know $s$ means the number of synthetic samples. But is it defined somewhere in the main text?} \fy{yes it is in the paragraph of ``Ensuring differential privacy''} 
   \STATE raw\_similarity $\gets \left[ \, \right]$
   
    % \STATE $\mathbf{p_{i}} \gets \mathrm{argMax}_{\mathbf{s_j} \in \{\mathbf{S_i}\}} \textbf{Score}(M, \mathbf{y_i}, \left[  \mathbf{s_j}; \mathbf{x_i}  \right])$ \label{alg:positive}

    \FOR{$j  \in \{1, \dots, p \}$}
    
    \STATE $S_{i, j} \gets sim(D_{syn, i}, D_{priv, j})$
    % cap raw similarity if noise multiplier > 0
    \IF{$\sigma >0$}
        \STATE $S_{i, j} = min(S_{i, j}, c)$
    \ENDIF

    \STATE raw\_similarity.append($S_{i, j}$)
    \ENDFOR

    \STATE raw\_similarity = sum(raw\_similarity)

    % add DP 
    \IF{$\sigma >0$}
        \STATE raw\_similarity += $\mathcal{N}(0, \sigma c \sqrt{s})$
    \ENDIF

    \STATE $R_{sim}\text{.append(}\frac{\text{raw\_similarity}}{p})$
\ENDFOR

\textbf{Return}: $R_{sim}$

\end{algorithmic}
\end{framed}
\caption{Procedure for computing similarity reward ($R_{sim}$) given a noise multiplier $\sigma$ for a batch of $s$ synthetic samples.}\label{alg:dp_voting}
\vspace{-1em}
\end{figure}
%%%%%%%%%%%%%%%%%

\paragraph{Ensuring differential privacy.} To ensure differential privacy, we add Gaussian noise to $R_{sim}$ following Aug-PE.  For a synthetic example $d$, we first obtain its similarity to each of the private documents $sim(d, D_{priv,i})$ and clip the similarity score to a threshold $c$. Given a noise multiplier $\sigma$ and a batch of synthetic samples of size $s$, we add Gaussian noise $\mathcal{N}(0, \sigma c \sqrt{s})$ to $\sum_{i} sim(d, D_{priv, i})$, i.e., the sum of similarity between the synthetic document and all private documents. Finally, we obtain the average similarity scores across all private documents. We provide the pseudocode of this procedure in  \autoref{alg:dp_voting}. 
The privacy analysis follows the Private Evolution framework \citep{pe1}, which is reproduced in \S \ref{app:privacy_analysis} for completeness.

\paragraph{Mitigating reward hacking.} Although optimizing for $R_{sim}$ can lead to synthetic documents similar to the private corpus defined by $sim$, relying only on $R_{sim}$ as the sole reward can lead to reward hacking \citep{Amodei2016ConcretePI, Pan2022TheEO}. Unlike prior RFT methods which focus on math and coding problems \citep{DeepSeekAI2025DeepSeekR1IR, Wei2025SWERLAL} where a verifiable reward (e.g. presence of the gold answer string) can be defined, generating documents that are similar to a set of documents is relatively open-ended and unconstrained. Moreover, text embedding models are known to have various biases \citep{coelho-etal-2024-dwell, Fayyaz2025CollapseOD} and could assign high similarity scores to dissimilar documents, especially when embedding \textit{long} documents. Thus, in addition to $R_{sim}$, we employ a reward $R_{prompt}$ that measures adherence to the input prompt to prevent reward hacking.\footnote{While RL algorithms such as PPO includes a KL penalty term to reduce divergence from the reference model, we find that increasing KL penalty does not resolve the issue in our preliminary experiments. Instead, it discourages the model from learning to generate samples closer to the private distribution.} We implement $R_{prompt}$ as a combination of LLM-as-a-judge and rule-based checks (\S \ref{sec:exps}).

To combine the two rewards, we use a threshold approach. We set the reward to 0 if $R_{prompt}$ is less than a threshold $\tau$ and otherwise return $R_{sim}$. The final reward is defined as:
\vspace{-0.3em}
\[
R = 
\begin{cases}
R_{\text{sim}} & \text{if } R_{\text{prompt}} > \tau \\%\text{ and } R_{\text{length}} = 1, \\
0 & \text{otherwise}
\end{cases}
\]

We train \method~ using Proximal Policy Optimization (PPO, \citet{Schulman2017ProximalPO}).

\section{Experiments}\label{sec:exps}

\subsection{Evaluation settings}
\paragraph{Datasets.} We evaluate DP-RFT and the baselines on four publicly available datasets that are treated as private datasets, covering different domains and document structures: 
\begin{itemize}[leftmargin=*]
\item \emph{Domain-specific document-level data:} \textbf{PubMed}, which contains abstracts of medical papers crawled by \citet{yu2023training} from 2023/08/01 to 2023/08/07; \textbf{BBC News article} from 2010 to 2017 released by \citet{Narayan2018DontGM}
\item \emph{Structured long-form data:} \textbf{WildChat}~\citep{zhao2024wildchat}, which contains chat history between a user and an LLM; collected between 2023 to 2024 and \textbf{QMSum}~\citep{zhong-etal-2021-qmsum} which contains meeting transcripts. 
\end{itemize}

The average number of tokens for PubMed, BBC article, WildChat and QMSum are $361$, $513$, $2{,}799$ and $2{,}857$ respectively. We include details about these datasets in \S \ref{app:dataset_details} in the appendix.

% We evaluate the generated synthetic documents  for both intrinsic and extrinsic evaluations.

\paragraph{Intrinsic evaluation: similarity with real data.} We report quantitative metrics including the average and maximal embedding similarity with the private corpus and the embedding distribution distance, i.e., the Fréchet Inception Distance \citep{heusel-fid}, following prior work~\citep{xie2024differentially}. As it is non-trivial to measure similarity of long documents, we also employ LLM-as-a-judge to evaluate pairwise similarity of two synthetic documents against a private document and conduct qualitative evaluation on document structure in \S \ref{sec:analysis}.

\paragraph{Extrinsic evaluation: downstream performance.} Aside from intrinsic similarity evaluation, the synthetic text should be helpful in downstream utility. We fine-tune a language model via next token prediction on the synthetic text generated and report the next token prediction accuracy on the private test set. Following prior work~\citep{yu2023training, xie2024differentially, popri2025}, we conduct experiments fine-tuning BERT\textsubscript{Small}~\citep{devlin2019bert} as a causal language model by modifying bidirectional attention to causal attention. As BERT\textsubscript{Small} is relatively small-scale, we also fine-tune and evaluate GPT-2~\citep{radford2019language}. %Following prior work~\citep{yu2023training, xie2024differentially, popri2025}, we fine-tune and evaluate BERT\textsubscript{Small}~\citep{devlin2019bert} as a causal language model by modifying bidirectional attention to causal attention.

\subsection{\method~and baseline settings}

We provide the implementation details for each component for \method~ as described in \S \ref{sec:method}.

$M$\textsubscript{gen} and $R_{sim}$: For all datasets, we use Qwen-2.5-3B-Instruct \citep{qwen25} as $M$\textsubscript{gen} and gte-Qwen-2-1.5B-Instruct~\citep{li2023towards} to measure $R_{sim}$. For WildChat and QMSum, we also measure the distributions of word-level jaccard similarity and word counts of all speech turn in a synthetic vs. private example, and use their KL divergence as two additional terms in $R_{sim}$, in order to encourage $M_{gen}$ to learn the structural properties of private corpus.

\textit{$R_{prompt}$}: We use gpt-4o~\citep{Hurst2024GPT4oSC} as the LLM-as-a-judge to evaluate how well the output adheres to the prompt. Concretely, given the input prompt $p$ and generated documents $d$, the model outputs a scalar rating from 1 to 10 on how well $d$ adheres to $p$. We include the exact prompt in \S \ref{app:implementation_details}. To evaluate adherence to the length instruction for the BBC article and PubMed, we employ a rule-based approach that checks the difference between the length of the generated documents and the specified length in the prompt. If the length difference is greater than a threshold $d$ or the LLM-as-a-judge output is less than a threshold $\tau$, we set the reward to 0. We set the threshold $\tau$ to 6 and $d$ as \{50, 100\} for PubMed and BBC articles respectively.  We conduct ablation studies on removing $R_prompt$ in \S ~\ref{subsec:ablation}.

\paragraph{Input prompts.} Following prior work~\citep{xie2024differentially}, we construct input prompts for each dataset by prompting an LM to generate keywords based on public knowledge. We design a multi-stage pipeline to ensure diversity of the prompts. For BBC articles, we first prompt the language model to generate a list of category of BBC articles. We then prompt the models to generate keywords for each categories. For PubMed, we use the writers released by \cite{xie2024differentially} and prompt the language model to generate 100 technical terms for each of the writer. The prompt template for each dataset in \autoref{tab:input_prompt_example} in the Appendix.

\paragraph{Baselines.} We consider several baseline methods which trains an LM with differential privacy. \textbf{Aug-PE}~\citep{xie2024differentially}, which steers generation from a language model to be more similar to the private data by iterative prompting;  \textbf{QWEN}: which prompts the backbone model to generate synthetic data; this represents the performance of the model before~\method{} training. We also compared to \textbf{DP-FT} which fine-tunes the model on the downstream task with DP-SGD \citep{yu2022differentially, li2022large}, though we note that \textbf{DP-FT} requires ingesting the private text as \textit{input} to train the model, and thus violating the \textit{eyes-off} setting. For all datasets, we follow the setup of \citet{xie2024differentially} and report \augpe{} results of 10 iterations. We use the same generation model and embedding model as~\method{} to ensure a fair comparison. We use the input prompt for~\method{} as \texttt{RANDOM\_API} and the fill-in-the-blank \texttt{VARIATION\_API}. We include exact prompts in \S \ref{app:baseline_settings} in the Appendix; as well as example private text and generated text in \S \ref{app:example_text}. We report results for training on a subset of private data (2,000) and compare with training on 2,000 data generated by ~\method{} and ~\augpe{}, except for QMSum, which only contains 700 examples in total. We generate synthetic text and keep the training data size consistent for all methods.

\paragraph{Privacy Setting.} We report the performance of different privacy budgets controlled by $\epsilon$. We train for 100 steps for PubMed and QMSum, and 200 steps for BBC articles and WildChat. We experiment with privacy budgets $\epsilon$ = \{1,2,4,$\infty$\}. The Gaussian noise multipliers being added to the DP votes are derived by the number of training steps and the size of the private corpus accordingly. The details are described in \S~\ref{app:privacy_analysis}. For DP-RFT, we use noise multiplier $\sigma$ = \{41.90, 22.14, 11.86, 0\} for PubMed, $\sigma$ = \{29.98, 16.45, 9.17, 0\} for QMSum, and $\sigma$ = \{52.50, 28.07, 15.23, 0\} for BBC articles and WildChat. We clip the maximum value of raw similarity score at 0.5 for BBC article and PubMed, and 0.8 for QMSum and WildChat.

\section{Results}
\begin{table*}%{r}{0.6\textwidth}
\caption{Evaluation on downstream model performance using Qwen-2.5-3B-Instruct as the synthetic data generator to finetune GPT2 and BERT\textsubscript{Small}. We note that \textbf{DP-FT} violates the \textit{eyes-off} constraint and color-coded the performance in grey. The highest performance among \textit{eyes-off} approaches is \textbf{bolded}. \method~ closes the gap between DP-FT and eyes-off approach such as ~\augpe{} on GPT-2, and outperform DP-FT when $\epsilon \neq \infty$ for BERT\textsubscript{Small}.} % {We report the result of using the full private corpus to train for DP-FT as well as using the same amount of generated data (2,000) to train.}
\begin{center}
% \scriptsize
\footnotesize
\begin{tabular}{@{}lllrrrrrrrrr@{}}
\toprule
\multirow{2}{*}{\textbf{Dataset}} & \multirow{2}{*}{\textbf{Method}} & \multirow{2}{*}{\textbf{Data Type}} & \multirow{2}{*}{\textbf{Size}} &  \multicolumn{4}{c}{\textbf{GPT-2 Next Token Accuracy ($\uparrow$)}}  &  \multicolumn{4}{c}{\textbf{BERT\textsubscript{Small} Next Token Accuracy ($\uparrow$)}} \\
& & & &  $\epsilon=\infty$ & $\epsilon=4$ & $\epsilon=2$ & $\epsilon=1$ &  $\epsilon=\infty$ & $\epsilon=4$ & $\epsilon=2$ & $\epsilon=1$  \\
\midrule
% \rowcolor{LightGrey} \multirow{4}{*}{\textbf{PubMed}} & DP-FT & {Private / full data} & 75,316 & \textbf{44.88} & \textbf{18.41} & \textbf{17.99} & \textbf{17.63}  \\
\rowcolor{LightGrey}\multirow{3}{*}{\textbf{PubMed}} & DP-FT & Private &  2,000 & {40.33} & {42.33} & {42.06} & {41.64} & {20.79} & 9 & 8 & 6 \\
& QWEN & Synthetic &  2,000 & 33.04  & 33.04 & 33.04 & 33.04  & 16.88  & {16.88} & {16.88} & \textbf{16.88}  \\
& \augpe & Synthetic &  2,000 &  33.91  & 34.10 & 33.57 & 34.55  & 13.96  & 14.41 & 14.16 & 13.71  \\ 
& \method  & Synthetic &  2,000 &  \textbf{38.31} & \textbf{37.37} & \textbf{37.19} & \textbf{36.75} &  \textbf{17.46} & \textbf{17.81} & \textbf{17.32} & {16.79} \\
\midrule 
% \rowcolor{LightGrey} \multirow{4}{*}{\textbf{BBC}} & DP-FT & {Private / full data }& 10,000 & \textbf{25.75} & \textbf{14.07} &	\textbf{13.71} & \underline{13.09} \\  
\rowcolor{LightGrey}\multirow{3}{*}{\textbf{BBC}}  &  DP-FT  & Private &  2,000 & {39.13} & {40.05} & {39.78} & {39.32} & {17.97} & 9.29 & 7.96 & 6.90 \\
& QWEN & Synthetic &  2,000 & 33.49 & \textbf{33.49} & \textbf{33.49} & {33.49} &  {13.07} & {13.07}  & \textbf{13.07}  & 13.07 \\
& \augpe & Synthetic &  2,000 &  30.97 & 31.78 & 31.54 & 32.25 &  11.63 & 10.78 & 10.03 & 10.00 \\ 
& \method  & Synthetic &  2,000 &  \textbf{33.57} & 32.81 & 32.84 & \textbf{33.92} &  \textbf{13.72} & \textbf{13.25} & {13.03} & \textbf{13.51} \\
% & \method  & Synthetic (2000) &  {23.07} & \textbf{22.38} & \textbf{22.00} & \textbf{22.91} \\
\midrule 
% \rowcolor{LightGrey}  \multirow{4}{*}{\textbf{WildChat}} & DP-FT & {Private / full data }& 10,000 & \textbf{28.78} &  {13.93} & {13.52} & {12.80}  \\ 
\rowcolor{LightGrey} \multirow{3}{*}{\textbf{WildChat}}  & DP-FT & Private &  2,000 & {51.00} & {47.95} & {47.59} & {47.06} & {19.62} & 8.28 & 7.37 & 5.30   \\
& QWEN & Synthetic &  2,000  & 33.00 & 33.00 & 33.00 & 33.00 & 5.22 & {5.22} & {5.22} & {5.22} \\
&  \augpe & Synthetic &  2,000 & 42.39 & 43.61 & 43.65 & 43.46 & 13.29 & {14.04} & {13.81} & {13.63}  \\ 
& \method & Synthetic &  2,000  & \textbf{45.27} & \textbf{45.42} & \textbf{45.30} & \textbf{45.03} & \textbf{13.93}& \textbf{14.08 }& \textbf{14.14} & \textbf{13.89} \\
\midrule
\rowcolor{LightGrey} \multirow{3}{*}{\textbf{QMSum}} & DP-FT & Private data & 700 & {47.00 }& {42.71} & {42.45} & {42.13}  & {32.82} & {11.52} & {9.88} & {7.96} \\ 
& QWEN & Synthetic &  700  & 24.00 & 24.00 & 24.00 & 24.00  & 3.86 & {3.86} & {3.86}  & {3.86}  \\
&  \augpe & Synthetic &  700 & 32.60 & 33.06 & 33.57 & \textbf{34.02} & 8.23 & 8.53 & 7.82 & 7.63 \\ 
& \method & Synthetic&  700 & \textbf{33.00} & \textbf{33.68} & \textbf{34.33} & 33.38 & \textbf{11.03} & \textbf{10.97} & \textbf{10.95} & \textbf{11.11} \\
\bottomrule
\end{tabular} 
\end{center}
\label{tab:downstream_results_gpt2}
\end{table*}

\subsection{Downstream evaluation}
\label{ssec:down_eval}

\autoref{tab:downstream_results_gpt2} reports the results on downstream performance for GPT-2 and BERT\textsubscript{Small}.

% \paragraph{\method{} outperform \augpe{} across different datasets and privacy budgets on both models.}  We observe up to 10\%, 4\%, 4\% and 2\% relative improvement compared to \augpe{} for GPT-2 for PubMed, BBC, WildChat and QMSum respectively. For BERT\textsubscript{Small}, we observe a similar trend with up to 22\%, 35\%, 2\% and 47 \% relative improvement on the four datasets respectively. This shows that \method{} is able to effectively shift the distribution of the LLM compared to \augpe{}, which iteratively prompts a frozen LLM. We also observe consistent improvement compared to the un-finetuned QWEN (except for BBC articles at $\epsilon=2$, where performance is comparable), demonstrating the effectiveness of training $M_{gen}$.

\paragraph{\method{} outperforms \augpe{} across datasets and privacy budgets.}
Across all datasets and privacy budgets, \method{} consistently achieves higher downstream accuracy than \augpe{} on both GPT-2 and BERT\textsubscript{Small}. The gains are particularly pronounced for BERT\textsubscript{Small}, with up to {22\%, 35\%, 2\%, and 47\%} relative improvement on PubMed, BBC, WildChat, and QMSum, respectively, and are also substantial for GPT-2, reaching up to {10\%, 4\%, 4\%, and 2\%} on these datasets. These improvements indicate that \method{} is able to effectively shift the distribution of the LLM compared to \augpe{}, which iteratively prompts a frozen LLM. In addition, \method{} consistently outperforms the un-finetuned QWEN baseline (except for BBC at $\epsilon=2$, where performance is comparable), further demonstrating the benefits of training $M_{\text{gen}}$.

% \paragraph{\method{} achieves better performance compared to DP-FT under tight privacy budgets and low data regime for BERT\textsubscript{Small}.}
% We note that while DP-FT has a looser privacy constraint (i.e. violating the \textit{eyes-off} setting), its performance degrades significantly with smaller $\epsilon$ value for BERT\textsubscript{Small}, underperforming ~\method{} for all datasets with $\epsilon \neq \inf$. In contrast, for GPT-2,  DP-FT consistently outperforms \method{} across different privacy budgets and datasets. We attribute this difference to the fact that BERT\textsubscript{Small} is converted into a casual language model during fine-tuning, which is a different set-up compared to pre-training. We conduct further experiment to scale-up the number of private data to the full set in \S \ref{app:downstream_settings}, observing improved downstream performance for DP-FT. Nevertheless, \method{} continues to outperform DP-FT for BERT\textsubscript{Small}  when the private data size is small (e.g. QMSum with only 700 datapoint in total) or under a tight privacy budget, highlighting its practical advantages in realistic privacy-constrained settings.

\paragraph{\method{} outperforms DP-FT under tight privacy budgets and low-data regimes for BERT\textsubscript{Small}.}
While DP-FT operates under a looser privacy constraint (i.e., violating the \textit{eyes-off} setting), its performance for BERT\textsubscript{Small} degrades substantially as the privacy budget decreases, underperforming \method{} across all datasets when $\epsilon \neq \infty$. In contrast, for GPT-2, DP-FT consistently outperforms \method{} across privacy budgets and datasets. We attribute this difference to the fact that BERT\textsubscript{Small} is converted into a causal language model during fine-tuning, which differs from its pre-training setup. To further investigate this effect, we scale up the amount of private data to the full dataset in \S\ref{app:downstream_settings}, observing improved downstream performance for DP-FT. Yet, \method{} continues to outperform DP-FT for BERT\textsubscript{Small} in low-data regimes (e.g., QMSum with only 700 examples) or under tight privacy budgets, highlighting its practical advantages in realistic privacy-constrained settings.

\subsection{Similarity evaluation}
\label{ssec:sim_eval}

% Table \ref{tab:similarity_results} reports similarity measurement with private data. Overall, ~\method{} demonstrates improved similarity with private documents compared to un-finetuned QWEN for both individual (mean and max similarity) and distributional metrics (FID), demonstrating the effectiveness of ~\method{}. ~\method{} also outperforms ~\augpe{} in most settings for both metrics, especially when privacy budget is small.  

\autoref{tab:similarity_results} reports similarity metrics between synthetic and private data. Overall, \method{} exhibits higher similarity to private documents than the unfine-tuned QWEN across both instance-level metrics (mean and maximum similarity) and distributional measures (FID), indicating that \method{} more effectively captures the underlying data distribution. In addition, \method{} outperforms \augpe{} for instance-level metrics in most settings, with the advantage becoming more pronounced under tighter privacy budgets.

% We also observe different trend for different datasets. For instance, we observe no performance differences between ~\augpe{} and \method{} in terms of embedding similarity for PubMed when $\epsilon = \infty$, compared to around 20\% improvement between for the other three datasets. This is potentially due to PubMed being closer to the distribution of the backbone model $M_{gen}$ and hence it is easier to steer. We measure the distance between the target distribution and $M_{gen}$ via perplexity of the private data on $M_{gen}$, finding the perplexity PubMed (6.75) to be lower than BBC article (9.11) and QMSum (10.14). While WildChat also has a low perplexity (4.04), its long-form, structural nature makes it hard to steer via iterative prompting. This is also reflected in FiD evaluation. We note that ~\augpe{} achieves better FiD for PubMed and WildChat. As \augpe{} is designed to optimize FiD, it is natural that it might achieve higher FiD, especially for tasks that are more familiar to the base model.

We observe dataset-dependent trends in similarity performance. For example, with $\epsilon = \infty$, \method{} and \augpe{} achieve comparable embedding similarity on PubMed, whereas \method{} yields approximately 20\% higher similarity on the other three datasets. We attribute this difference to PubMed being closer to the distribution of the backbone generator $M_{\text{gen}}$, making it easier to steer via prompting. To quantify this, we measure the distance between the target distribution and $M_{\text{gen}}$ using perplexity of the private data under $M_{\text{gen}}$, finding that PubMed has lower perplexity (6.75) than BBC (9.11) and QMSum (10.14). Although WildChat also exhibits low perplexity (4.04) as it contains model-generated texts, its long-form and structured nature makes it more difficult to steer through iterative prompting. %We note that \augpe{} achieves lower FiD on PubMed and WildChat, given that \augpe{} explicitly optimizes FiD and thus may perform better on distributions that are more familiar to the base model.
We note that \augpe{} achieves lower FiD on PubMed and WildChat, which may be attributed to the fact that \augpe{} explicitly optimizes FiD and can therefore perform better on distributions that are more familiar to the base model.
 
 To complement embedding similarity, we  conduct an LLM-as-a-judge pairwise similarity evaluation against private document on use of word and tone, finding that ~\method{} is consistently preferred over ~\augpe{} across all datasets. We describe the setting and results in details in \S \ref{app:llm_as_a_judge} in the Appendix.

\begin{table*}%{r}{0.6\textwidth}
\caption{Similarity evaluation with the private corpus using Qwen-2.5-3B-Instruct as the synthetic data generator. \method{} generates documents that are more similar to the private dataset than baseline methods as measured by embedding similarity.}
\begin{center}
% \scriptsize
\footnotesize
\begin{tabular}{@{}llrrrrrrrr@{}}
\toprule
\multirow{2}{*}{\textbf{Dataset}} & \multirow{2}{*}{\textbf{Method}} &  \multicolumn{4}{c}{\textbf{Mean/Max embedding similarity ($\uparrow$)}} & \multicolumn{4}{c}{\textbf{FID ($\downarrow$)}} \\
& &  $\epsilon=\infty$ & $\epsilon=4$ & $\epsilon=2$ & $\epsilon=1$ & $\epsilon=\infty$ & $\epsilon=4$ & $\epsilon=2$ & $\epsilon=1$ \\
\midrule
\rowcolor{LightGrey} \multirow{3}{*}{\textbf{PubMed}} & Private & {0.54/0.88} & - & - & - & {0.06} & - & - & - \\ 
& QWEN & 0.44/0.73 & - & - & - & 0.50 & - & - & - \\ 
& \augpe & \textbf{0.48/0.77} & \textbf{0.49/0.79} & 0.46/0.74 & 0.43/0.70 & \textbf{0.39} & \textbf{0.39} & \textbf{0.41} & \textbf{0.46} \\ 
& \method &  0.47/0.70 & \textbf{0.49}/0.71 & \textbf{0.51/0.73} & \textbf{0.52/0.74} & 0.46 & 0.48 & 0.45 & 0.47 \\ 
\midrule
\rowcolor{LightGrey}  \multirow{3}{*}{\textbf{BBC}}  & Private & {0.43/0.78} & - & - & - & {0.09} & - & - & - \\ 
& QWEN & 0.27/0.52 & - & - & - & 0.64 & - & - & - \\ 
& \augpe & 0.30/0.55 & 0.21/0.45 & 0.20/0.43 & 0.19/0.42 & \textbf{0.49} & 0.68 & 0.78 & 0.79 \\ 
& \method & \textbf{0.35/0.59} & \textbf{0.33/0.57} & \textbf{0.34/0.57} & \textbf{0.35/0.58} & 0.55 & \textbf{0.57} & \textbf{0.57} & \textbf{ 0.52} \\
\midrule 
\rowcolor{LightGrey} \multirow{3}{*}{\textbf{WildChat}}  & Private &  {0.28/0.67} & - & - & - & {0.07} & - & - & - \\ 
& QWEN & 0.27/0.64  & - & - & -  & 1.06 & - & - & -  \\ 
& \augpe & 0.24/0.57 & 0.25/0.57 & 0.24/0.57 & 0.23/0.54 & \textbf{0.39} & \textbf{0.32} & \textbf{0.39} & \textbf{0.38} \\ 
& \method & \textbf{0.30/0.78} & \textbf{0.31/0.77} & \textbf{0.31/0.78} & 0.\textbf{30/0.77} & 0.74 & 0.66 & 0.71 & 0.71\\ 
\midrule
\rowcolor{LightGrey} \multirow{3}{*}{\textbf{QMSum}} &  Private & {0.52/0.84}  & - & - & - & {0.24} & - & - & - \\  
& QWEN &  0.32/0.48 & - & - & - & 1.11  & - & - & - \\ 
& \augpe & 0.34/0.54 & 0.30/0.49 & 0.31/0.49  & 0.30/0.48 & \textbf{0.70} & 0.79 & 0.80 & \textbf{0.82} \\ 
& \method & \textbf{0.37/0.62} & \textbf{0.38/0.65} & \textbf{0.38/0.64} & \textbf{0.38/0.65} & 0.80 & 0.79 & \textbf{0.75} & 0.84  \\ 
\bottomrule
\end{tabular} 
\end{center}
\label{tab:similarity_results}
\end{table*}

\section{Analysis and ablation}\label{sec:analysis}

\begin{figure*}[t]
    \centering

    % --- Top figure ---
    \begin{subfigure}{\linewidth}
        \centering
        \includegraphics[width=0.7\linewidth]{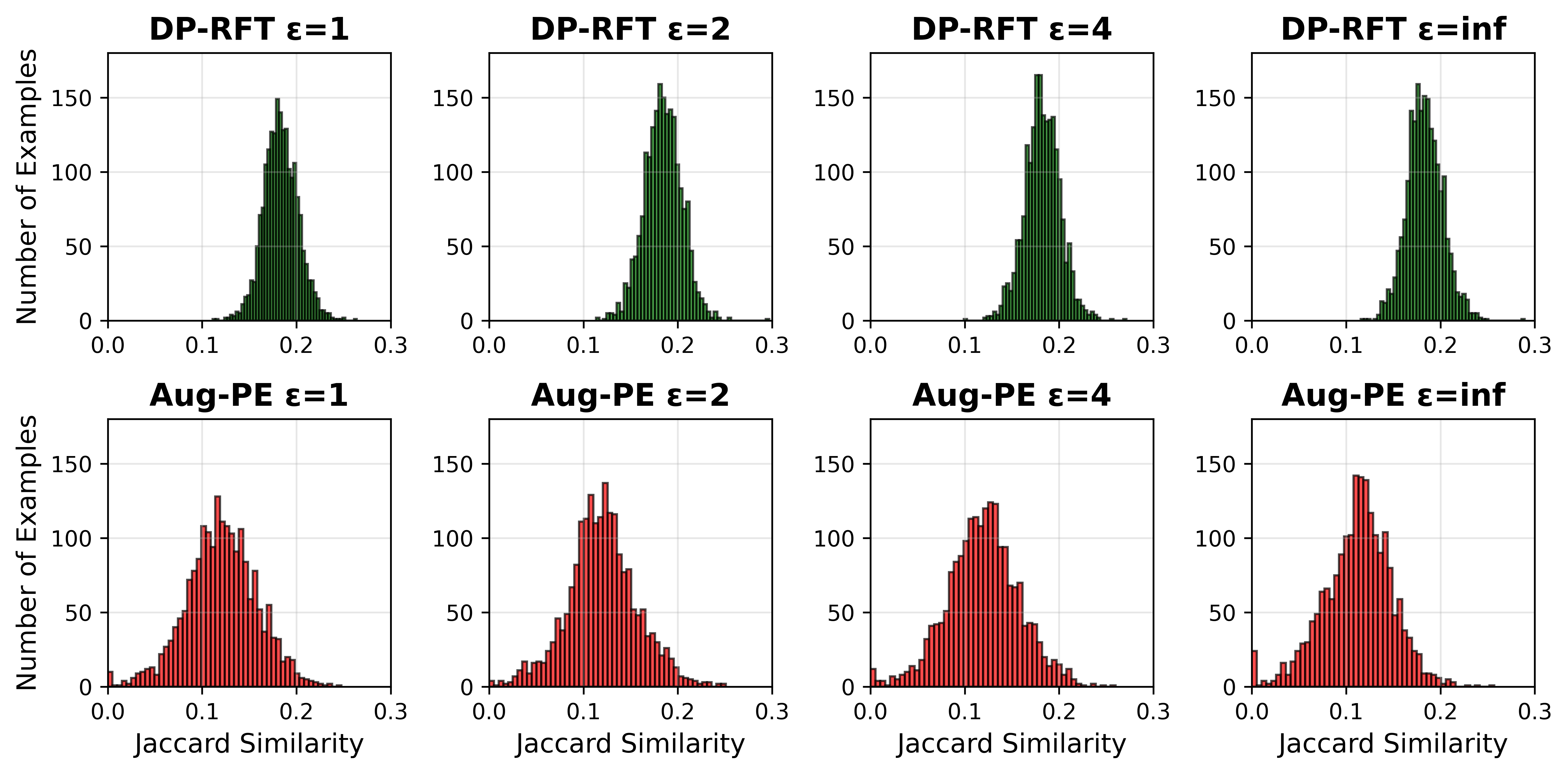}
        % \caption{Histogram of word-level maximum Jaccard similarity of synthetic WildChat examples against all private examples, comparing DP-RFT vs.\ Aug-PE under different privacy budgets.}
        % \label{fig:jaccard}
    \end{subfigure}

    \vspace{0.6em}

    % --- Bottom figure ---
    \begin{subfigure}{\linewidth}
        \centering
        \includegraphics[width=0.8\linewidth]{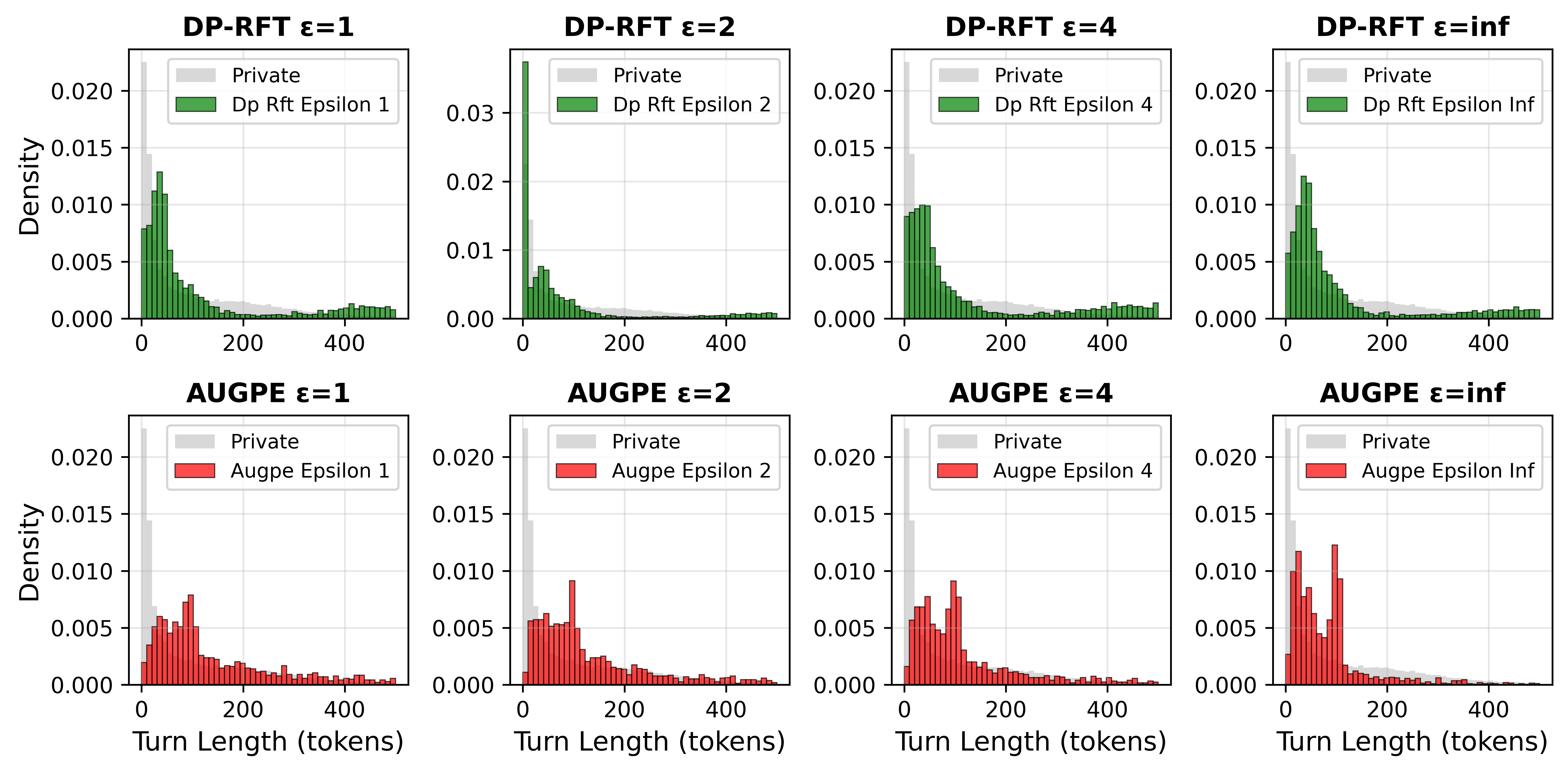}
        % \caption{Distribution of per-turn word counts from synthetic WildChat chatlogs generated by DP-RFT vs.\ Aug-PE under different privacy budgets ($\epsilon=\{1,2,3,\infty\}$). Real (private) WildChat examples are shown in gray.}
        % \label{fig:turn-dist}
    \end{subfigure}

    \caption{
    Analysis of structural similarity on WildChat. Top: Histogram of word-level maximum Jaccard similarity of synthetic WildChat examples against all private examples, comparing DP-RFT vs.\ Aug-PE under different privacy budgets. Bottom: Distribution of per-turn word counts from synthetic WildChat chatlogs generated by DP-RFT vs.\ Aug-PE under different privacy budgets ($\epsilon=\{1,2,3,\infty\}$). Real (private) WildChat examples are shown in gray.
    }
    \label{fig:wildchat-analysis}
\end{figure*}

\subsection{Qualitative Analysis on the Synthetic Examples}\label{ssec:wildchat-analysis}
To understand the effect of DP-RFT training and the properties the generated synthetic data, we analyze the distributional similarity between synthetic vs. real on private WildChat conversations. 

% \paragraph{Lexical Similarity.} We plot the distribution of Jaccard similarity between each synthetic example and its \textit{most} similar private example in the private corpus. The results are shown in \autoref{fig:wildchat-analysis}. We see that across different privacy budgets, examples generated by DP-RFT demonstrate higher lexical overlap to the private distribution compared to ~\augpe. Interestingly, \augpe~ distribution is more spread out, potentially indicating that Aug-PE outputs are more diverse, despite lower fidelity in terms of lexical similarity. This aligns with our FiD evaluation in \S \ref{ssec:sim_eval}, showing that ~\augpe{} achieves better FiD compared to ~\method{} despite worse mean embedding similarity. Improvement on ~\method{} which considers diversity of the generated samples is a promising future direction.

\paragraph{Lexical Similarity.}
We analyze the distribution of Jaccard similarity between each synthetic example and its \textit{most} similar private example, as shown in \autoref{fig:wildchat-analysis}. Across privacy budgets, examples generated by DP-RFT exhibit higher lexical overlap with the private data than those produced by \augpe{}. In contrast, the distribution of examples from \augpe{} is more spread out, potentially indicating that \augpe{} outputs are more diverse, despite lower fidelity in terms of lexical similarity. This observation is consistent with the FiD results in \S\ref{ssec:sim_eval}, where \augpe{} achieves lower FiD despite worse average embedding similarity. Incorporating more explicit diversity control into \method{} is a promising direction for future work.

% \paragraph{Structural Similarity.} For structured data, both content similarity and structural similarity of synthetic examples are important.
% \autoref{fig:wildchat-analysis} shows the distribution of per-turn word counts from the synthetic vs. real chatlogs in WildChat across different privacy budgets.  Overall, we observe that compared to ~\augpe{}, the generated synthetic data from~\method{} models better resembles the private distribution across all privacy budget settings. Yet interestingly, ~\augpe{} is able to pick up the bi-modal pattern of the distribution, where most of the responses are short (e.g. user turn), and other responses tend to be longer. A potential explanation here is that Aug-PE uses the generate-then-select strategy to sample the entire group of samples at each iteration, while ~\method's optimization steps happen for mini-batches of smaller sample size. Future work can explore the impact of roll-out batch sizes during ~\method{}'s training.  %We see that the real distribution resembles a bi-modal pattern, where most of the responses are short (e.g. user turn), and other responses tend to be longer.

\paragraph{Structural Similarity.}
For structured data, both content and structural similarity of synthetic examples are important. \autoref{fig:wildchat-analysis} compares the per-turn word-count distributions of synthetic and real WildChat conversations across privacy budgets. Overall, synthetic data generated by \method{} more closely matches the private distribution than \augpe{} across all privacy settings. Interestingly, however, \augpe{} better captures the bi-modal structure of the distribution, reflecting the mixture of short user turns and longer assistant responses. One possible explanation is that Aug-PE employs a generate-then-select strategy over a large candidate set at each iteration~\citep{xie2024differentially}, whereas \method{} optimizes over smaller mini-batches of rollouts. Future work can examine the effect of rollout batch size in \method{} training more closely.

%  \begin{figure*}[h!]
%     \centering
%     \includegraphics[width=0.7\linewidth, trim={0 0 0 0},clip]{figs/wildchat_jaccard_similarity_plot.png}
%     \caption{The histogram of word-level maximum Jaccard similarity of synthetic WildChat examples against all private examples, comparing DP-RFT vs. Aug-PE under different privacy budgets. Higher jaccard similarity indicates higher lexical overlap between synthetic and private examples. }
%     \label{fig:jaccard}
% \end{figure*}

%  \begin{figure*}[h!]
%     \centering
%     \includegraphics[width=0.8\linewidth, trim={0 0 0 0},clip]{figs/wildchat_turn_length_plot.png}
%     \caption{The distribution of word counts per-turn from synthetic WildChat chatlogs generated by DP-RFT vs. Aug-PE under different privacy budgets ($\epsilon=\{1, 2, 3, \infty\}$). For comparison, the distribution of word counts in real (``private'') WildChat examples is shown in gray for each plot. }
%     \label{fig:turn-dist}
% \end{figure*}

\subsection{Ablation study}\label{subsec:ablation}

We propose to include $R_{prompt}$ to mitigate reward hacking by over-optimizing on $R_{sim}$ (\S \ref{subsec:reward_design}). We conduct ablation study which removes $R_{prompt}$ from the reward on PubMed  with $\epsilon=\infty$ and report evaluation on both downstream performance and similarity measure. Results are reported in \autoref{tab:reward_ablation}. We see that while model trained without $R_{prompt}$ (using $R_{sim}$ as the reward) achieves higher embedding similarity with the private corpus, the downstream performance is notably worse across the two models. This suggests that solely optimizing for embedding similarity could harm performance. We further conduct ablation study on varying $R_{sim}$ and $M_{gen}$ in \S \ref{app:more_ablations}.

\begin{table}%{r}{0.5\textwidth}\vspace{-1em}
\caption{Ablating $R_{sim}$ for QMSum when $\epsilon=\infty$. Removing $R_{prompt}$ leads to better embedding similarity but results in worse downstream performance.}\vspace{-0.5em}
\small
\begin{center}
\begin{tabular}{@{}lccc@{}}
\toprule
\multirow{2}{*}{\textbf{Method}}& \textbf{Acc ($\uparrow$)} &   \multicolumn{2}{c}{\textbf{Embedding sim}} \\
& \textbf{GPT-2 / BERT\textsubscript{Small}} & \textbf{mean/max($\uparrow$)} & \textbf{FID ($\downarrow$)}  \\
\midrule
\rowcolor{LightGrey} \textbf{DP-FT} &  42.33/20.97 & 0.54/0.88 & 0.06 \\ 
\textbf{Aug-PE} & 33.91/13.96 & 0.48/0.77 & 0.39 \\ 
\textbf{DP-RFT} & 38.31/17.46 & 0.47/0.70 & 0.46 \\ 
 - $R_{prompt}$ & 36.85/16.31 & 0.56/0.78 & 0.45 \\ 
\bottomrule
\end{tabular}
\end{center}
\label{tab:reward_ablation}
\end{table}

% \subsection{Computational overhead.} Computational costs for generating the synthetic data is an important axis to consider for downstream application. We benchmark the computational costs by ~\method{} and ~\augpe{} with wall-clock GPU hours needed for each method in our experiment settings on WildChat. \textbf{\method} consists of two phases: \textbf{(1) Training}: 200 steps complete in about 5 hours with 8*A100 GPUs, totaling 40 GPU hours. \textbf{(2) Inference}: Generating 2000 samples take 1 A100 hours with vLLM~\citep{kwon2023efficient} inference. \textbf{~\augpe}: generating 2000 samples for each iterations takes 10 GPU hours, totaling to 100 GPU hours for 10 iterations\footnote{We use the official implementation in \url{https://github.com/microsoft/DPSDA}.}. While Aug-PE does not involve model training, the iteration requires substantial computational cost, especially for generating longer synthetic text, below, we benchmark the time required to generate synthetic data by DP-RFT and Aug-PE for WildChat. Note that DP-RFT’s training is a one-time cost.

\subsection{Computational Overhead.}
Computational cost is an important consideration for practical deployment of synthetic data generation methods. We benchmark the wall-clock GPU hours required by \method{} and \augpe{} under our WildChat experimental setting. \method{} consists of two phases: \textbf{(1) Training}, where 200 optimization steps require approximately 5 hours on 8 A100 GPUs (40 GPU hours total), and \textbf{(2) Inference}, where generating 2{,}000 samples requires roughly 1 A100 GPU hour using vLLM~\citep{kwon2023efficient}. In contrast, \augpe{} does not involve model training but incurs substantial cost during iterative generation: producing 2{,}000 samples per iteration requires approximately 10 GPU hours, resulting in a total of 100 GPU hours over 10 iterations\footnote{We use the official implementation at \url{https://github.com/microsoft/DPSDA}.}. While \augpe{} avoids a training phase, its iterative procedure is computationally expensive, particularly for long-form text generation. In comparison, the training cost of \method{} is a one-time expense, after which synthetic data can be generated efficiently.

\section{Conclusion}
 % We introduce ~\method{}, a method to train language model to generate synthetic data similar to a private corpus \textit{without} ingesting any private document as model input. The key idea of our method is to steer the model with reinforcement learning guided by a similarity measure with the private document as reward. We conduct comprehensive evaluation on four datasets of our method against baseline methods such as ~\augpe{}, showing the effectiveness of the generated data with both similarity and downstream utility.

We introduce \method{}, a reinforcement learning method for training language models to generate synthetic data aligned with a private corpus \textit{without} ingesting individual private examples. The key idea of our method is to steer the model with reinforcement learning guided by a similarity measure with the private document as reward, while respecting the eyes-off constraint. Through extensive experiments on four datasets spanning diverse domains and structures, we show that \method{} consistently improves both fidelity and downstream utility over prior eyes-off baselines such as \augpe{}, and narrows the gap with DP finetuning methods under tight privacy budgets.


% Authors are kindly asked to make their submissions as accessible as possible
% for everyone including people with disabilities and sensory or neurological
% differences. Tips of how to achieve this and what to pay attention to will be
% provided on the conference website \url{http://icml.cc/}.

% \section*{Software and Data}

% If a paper is accepted, we strongly encourage the publication of software and
% data with the camera-ready version of the paper whenever appropriate. This can
% be done by including a URL in the camera-ready copy. However, \textbf{do not}
% include URLs that reveal your institution or identity in your submission for
% review. Instead, provide an anonymous URL or upload the material as
% ``Supplementary Material'' into the OpenReview reviewing system. Note that
% reviewers are not required to look at this material when writing their review.

% % Acknowledgements should only appear in the accepted version.
% \section*{Acknowledgements}

% \textbf{Do not} include acknowledgements in the initial version of the paper

% If a paper is accepted, the final camera-ready version can (and usually should)
% include acknowledgements.  Such acknowledgements should be placed at the end of
% the section, in an unnumbered section that does not count towards the paper
% page limit. Typically, this will include thanks to reviewers who gave useful
% comments, to colleagues who contributed to the ideas, and to funding agencies
% and corporate sponsors that provided financial support.

\section*{Impact Statement}
This work advances machine learning by exploring a reinforcement learning–based approach to differentially private synthetic data generation without eyes-on access to private examples. By using DP-protected similarity signals as training feedback, ~\method{} enables large language models to generate synthetic texts better aligned with private data distributions while respecting formal privacy constraints. This reduces the need for direct exposure to sensitive data and may help lower privacy risks in applications where data sharing is restricted. More broadly, our results suggest a path toward more practical and responsible use of LLMs in privacy-sensitive domains.


\bibliography{example_paper}
\bibliographystyle{icml2026}

%%%%%%%%%%%%%%%%%%%%%%%%%%%%%%%%%%%%%%%%%%%%%%%%%%%%%%%%%%%%%%%%%%%%%%%%%%%%%%%
%%%%%%%%%%%%%%%%%%%%%%%%%%%%%%%%%%%%%%%%%%%%%%%%%%%%%%%%%%%%%%%%%%%%%%%%%%%%%%%
% APPENDIX
%%%%%%%%%%%%%%%%%%%%%%%%%%%%%%%%%%%%%%%%%%%%%%%%%%%%%%%%%%%%%%%%%%%%%%%%%%%%%%%
%%%%%%%%%%%%%%%%%%%%%%%%%%%%%%%%%%%%%%%%%%%%%%%%%%%%%%%%%%%%%%%%%%%%%%%%%%%%%%%
\newpage
\appendix
\onecolumn

\section{Large Language Model Usage}
We use services powered by large language models (through ChatGPT and Grammarly) to fix grammatical errors and polish the writing in this submission. We do not use LLM-aid for other aspects of the paper writing.   

\section{Reproducibility statement}
We include implementation details for reproducing ~\method{} in Section \ref{app:implementation_details} and for reproducing baseline methods in Section \ref{app:baseline_settings}. Implementation details for downstream evaluation is documented at \ref{app:downstream_settings}. 

\section{Implementation details of DP-RFT}\label{app:implementation_details}

\paragraph{Training} We train both the actor and critic models using PPO algorithm on a single node with 8 A100 GPUs. We set the batch size to 128, with two samples per 64 samples each. We set the max completion tokens to 512, 1024, 2,048 and 2,048 for PubMed abstract, BBC article, QMSum and WildChat respectively. We train the model with the \texttt{veRL} library.

\paragraph{Input prompts} For BBC articles, we sample length from a Gaussian distribution with mean of 400 words and standard deviation of 100 words within the range of [100, 900] words (inclusive). We round the number of words to the closest 100. For PubMed, we sample uniformly from [100, 400] words and round the number to the closest 50. We include the example input prompt for each dataset in  \autoref{tab:input_prompt_example}. We include prompt used to construct the keywords in the input prompt in  \autoref{tab:keyword_generation_prompt}.

\paragraph{Reward model} We include the prompt used for obtaining $R_{prompt}$ in Table \ref{tab:llm_as_judge_prompt}.

\section{Privacy Analysis}
\label{app:privacy_analysis}
The privacy analysis of \method{} follows the Private Evolution framework \citep{pe1}, with modifications only in the sensitivity analysis of the mechanism.

Firstly, we analyze the privacy cost of the procedure in Figure~\ref{alg:dp_voting}. Each private sample contributes a vector of size $s$. By the clipping operation, each entry of the vector has absolute value at most $c$, so its $\ell_2$ norm is bounded by $c\sqrt{s}$. Consequently, adding or removing a single private sample changes $R_{\text{sim}}$ by at most $c\sqrt{s}$ in $\ell_2$ norm. Adding i.i.d.\ Gaussian noise with standard deviation $\sigma c \sqrt{s}$ then corresponds to the standard Gaussian mechanism \citep{dwork2014algorithmic} with noise multiplier $\sigma$.

The overall \method{} algorithm performs $T$ iterations of Figure~\ref{alg:dp_voting}, which can be viewed as $T$ adaptive compositions of the Gaussian mechanism. By Corollary~3.3 in \citet{dong2022gaussian},\footnote{See the arXiv version: \url{https://arxiv.org/pdf/1905.02383}.} this is equivalent to a single Gaussian mechanism with effective noise multiplier $\sigma/\sqrt{T}$. Therefore, we can apply the tight bounds for Gaussian mechanisms from \citet{balle2018improving} to analyze the privacy cost. In particular, given any two of $\epsilon$, $\delta$, and $\sigma/\sqrt{T}$, their analysis allows us to solve for the third. For example, for a target privacy budget $(\epsilon,\delta)$, one can compute the required noise multiplier $\sigma/\sqrt{T}$, which in turn specifies the amount of noise $\sigma c\sqrt{s}$ to add to $R_{\text{sim}}$.

The above privacy computation procedure is implemented in the open-source Private Evolution library.\footnote{\url{https://github.com/microsoft/DPSDA/blob/main/pe/dp/gaussian.py}}

\begin{table*}[ht!]
\center
\caption{Prompt used for constructing input prompts, as described in Section \ref{sec:method}.}
\footnotesize
\begin{tabular}{p{2cm}p{10cm}}
\toprule
\textbf{Stage} & \textbf{Prompt} \\ 
\midrule
\multicolumn{2}{c}{\textit{BBC article}} \\ 
Stage 1 & You will generate a list of 500 keywords for a subcategory, which will be used to write a BBC news articles for events between 2010 and 2017.
The keywords should be relevant to the subcategory, and they should be diverse and do not repeat each other.
The keywords can be relevant entities, names, events, or any other relevant terms.
The subcategory is \{category\_str\}. Number the keywords and return the list of keywords separated by newline. Do not return anything else. \\ 
Stage 2 & You will be given a category and a keyword related to that category that will be used to write BBC news articles for events between 2010 and 2017.
You will augment the keyword with 4 other keywords, that are relevant to the category and the keyword.
The sets of keywords should be relevant to the subcategory so that they can be used to write a BBC news articles.
The category is \{category\_str\} and the keyword is \{keyword\_str\}. Return the five keywords separated by comma, do not return anything else. \\
\midrule 
\multicolumn{2}{c}{\textit{PubMed}} \\ 
Stage 1 & Suppose that you are a \{writer\}. Please provide a list of 100 technical terms that are introduced by a PubMed journal article.
The keywords should be diverse and do not repeat each other.
Number the keywords and return the list of keywords separated by newline. Do not return anything else. \\ 
\bottomrule
\end{tabular}
\label{tab:keyword_generation_prompt}
\end{table*}

\begin{table*}[ht!]
\center
\caption{Example input prompts for each dataset.}
\footnotesize
\begin{tabular}{p{2cm}p{10cm}}
\toprule
\textbf{Dataset} & \textbf{Input prompt} \\ 
\midrule
\textbf{BBC article} & You are an AI assistant that acts as a journalist and write an article that mimics the style of BBC news article. Please write a BBC news article using the below keywords: Sustainable Development Goals, climate change, renewable energy, global partnership, poverty eradication. 

News Article: \\ 
\midrule
\textbf{PubMed} &  Suppose that you are a Clinical Researcher. Please provide an example of an abstract for a medical research paper on the topic of Platelet aggregation.  The abstract should contain around 250 words. Abstract: \\
\midrule
\textbf{QMSum} & You are given a topic and a list of pariticipants. Your task is to generate a meeting transcript where the participants discuss around the topic. 
The transcript should contain utterances, where each utterance is of format "Participant: Content". Separate each utterance with \texttt{\textbackslash n}. 

\#\#\# Participants: Software Engineer, Product Manager, Quality Assurance, UX Designer

\#\#\# Topic: New mobile app feature brainstorming

\#\#\# Transcript: \\
\midrule
\textbf{WildChat} & Generate a realistic and engaging conversation between a user and ChatGPT.
The user initiates the conversation with a request related to **Health \& Wellness**. \\
& The task should involve **Research**. \\
& The tone of the conversation should be **Inquisitive**, and the assistant’s response style should be **Narrative**. \\
& You should output the conversation in the following format. \\
& Each participant's dialogue should be on a new line. \\
& Do not include any other text or formatting. \\
& User: [User's message] \\
& Assistant: [Assistant's message] \\
& User: [User's message] \\
&... \\

& The conversations should ideally contain multiple turns. Only output the conversation in the format above. Do NOT output anything else.",
  \\ 
\bottomrule
\end{tabular}
\label{tab:input_prompt_example}
\end{table*}

\begin{table*}[ht!]
\center
\caption{Prompt used for obtaining $R_{prompt}$.}
\footnotesize
\begin{tabular}{p{13cm}}
\toprule
\textbf{Prompts} \\ 
\midrule
\# Instruction \newline
Please act as an impartial judge and evaluate the quality of the response provided by an AI assistant to the user instruction displayed below. Your evaluation should consider factors such as the helpfulness, relevance, depth, creativity, and level of detail of the response. Be as objective as possible. Only respond the rating number from 1 to 10. Do not include any other text or explanation in your response. Do not respond with anything else after [Rating], just the score by itself. \newline
[Instruction] \newline
\texttt{\{instruction\}} \newline 
[The Start of Assistant's Response] \newline
\texttt{\{response\}} \newline
[The End of Assistant's Response] \newline
[Rating] \\
\bottomrule
\end{tabular}
\label{tab:llm_as_judge_prompt}
\end{table*}

\begin{table*}[ht!]
\center
\caption{Prompt for LLM-as-a-judge for similarity analysis.}
\footnotesize
\begin{tabular}{p{13cm}}
\toprule
\textbf{Prompts} \\ 
\midrule
\# Instruction \newline
Please act as an impartial judge and evaluate the stylistic similarity of the responses provided by two AI assistants compared to the reference response. You should consider the elements such as use of words, style and tone. Do not consider the content match as the subject matter of the reference answer is different. Begin your evaluation by comparing both assistants' answers with the reference answer. Identify the similarity and differences. Avoid any position biases and ensure that the order in which the responses were presented does not influence your decision.  Do not allow the length of the responses to influence your evaluation. Do not favor certain names of the assistants. Be as objective as possible. After providing your explanation, output your final verdict by strictly following this format: [[A]] if assistant A is better, [[B]] if assistant B is better, and [[C]] for a tie.\newline
[Reference answer]\newline
\texttt{\{reference\_answer\}} \newline
[The Start of Assistant A's Answer]\newline
\texttt{\{assistant a answer\}}\newline
[The End of Assistant A's Answer]\newline
[The Start of Assistant B's Answer]\newline
\texttt{\{assistant b answer\}}\newline
[The End of Assistant B's Answer] \\

\bottomrule
\end{tabular}
\label{tab:prompt_used_for_similarity_analysis}
\end{table*}

\begin{table*}[ht!]
\center
\caption{\texttt{VARIATION\_API} prompt for the PubMed dataset.}
\footnotesize
\begin{tabular}{p{13cm}}
\toprule
You are required to fill in the blanks with more details for the input medical abstract in a professional tone. If there is no blanks, please output the original medical abstract.
Please fill in the blanks in the following sentences to write an abstract of a medical research paper: \{masked\_sample\} and your answer MUST be exactly \{word\_count\} words. \\
\bottomrule
\end{tabular}
\label{tab:augpe_prompt}
\end{table*}

\section{Experiment details}

\subsection{Datasets}\label{app:dataset_details}

\begin{table}
\caption{Data statistics of the private data. We report the average number of tokens measured by Qwen-2.5-Instruct.}
\small
\begin{center}
\begin{tabular}{lccccc}
\toprule
  {\textbf{Dataset}} &  {\textbf{Average \# of tokens}}  &  {\textbf{Private corpus size}} &  \textbf{Private test set size} \\
  \midrule
  \textbf{PubMed} & 361 & 75,316 & 4,453 \\
  \textbf{BBC article} & 513 & 10,000 & 11,057  \\
  \textbf{WildChat} & 2,799 & 10,000 & 8,891 \\
  \textbf{QMSum} & 2,857 & 725 & 296 \\ 
\bottomrule
\end{tabular} 
\end{center}
\label{tab:datasets_stats}
\end{table}

We report dataset statistics for all four datasets in \autoref{tab:datasets_stats}.

\subsubsection{Baseline details}\label{app:baseline_settings}
\paragraph{Aug-PE} We use the input prompt for ~\method{} as \texttt{RANDOM\_API} for ~\augpe. We include samples of input prompts in \autoref{tab:input_prompt_example}. For \texttt{VARIATION\_API}, we use the \texttt{fill-in-the-blank} following previous work~\cite{xie2024differentially}. We include the prompt in \autoref{tab:augpe_prompt}. We set the max completion tokens the same as ~\method{} and the blank probability to 0.6.

% \paragraph{Ablating the similarity function.}

% To understand the effect of the Jaccard similarity and turn length KL divergence reward terms we use for WildChat and QMSum's $R_{sim}$ design, we study the effect of different combination of reward terms for the DP-RFT model. The results are shown in Table \ref{tab:reward_ablation}.  We observe that the choice of reward leads to small yet not significant changes both in terms of the downstream evaluation and intrinsic evaluation. However, we notice that qualitatively having the two rewards will lead to the model better learning the structural properties from the private distribution (i.e. meeting transcripts and chat logs), as we demonstrate in Section \ref{ssec:wildchat-analysis}. 

% \begin{table}%{r}{0.5\textwidth}\vspace{-1em}
% \caption{Ablating $R_{sim}$ for QMSum when epsilon=4.}\vspace{-0.5em}
% \small
% \begin{center}
% \begin{tabular}{@{}lccc@{}}
% \toprule
% \textbf{$R_{sim}$} & \textbf{BERT Accuracy ($\uparrow$)} &   \textbf{Mean/max similarity ($\uparrow$)} &  \textbf{FID ($\downarrow$)}\\ \midrule
% \method  & 10.97 & 0.38/0.65 & 0.79 \\ 
% % embedding & 11.32 & 0.36/0.62 & 0.73 \\
% \method - jaccard - KL & 11.32 & 0.36/0.62 & 0.73 \\
% % embedding  + jaccard + $R_{prompt}$ & 11.54 & 0.38/0.65 & 0.75 \\
% \method  - KL & 11.54 & 0.38/0.65 & 0.75 \\
% % embedding + jaccard + KL & 10.99 & 0.29/0.52 & 0.76 \\ 
% \method - $R_{prompt}$ & 10.99 & 0.29/0.52 & 0.76 \\ 
 
% \bottomrule
% \end{tabular} 
% \end{center}
% \label{tab:reward_ablation}
% \end{table}

\section{Ablation study}\label{app:more_ablations}

\begin{table}
\caption{Upper: Ablating embedding models used to estimate similarity with the private corpus $R_{sim}$ on the PubMed dataset with QWEN-2.5-3B-Instruct as $M_{gen}$. Bottom: Ablating language models $M_{gen}$ on BBC article with $R_{sim}$ as gte-1.5B.}
\small
\begin{center}%
\begin{tabular}{@{}llcc@{}}
\toprule
\multirow{2}{*}{\textbf{$R_{sim}$}} & \multirow{2}{*}{\textbf{Method}} & \multicolumn{2}{c}{\textbf{BERT accuracy}} \\
 & & $\epsilon=\infty$ & $\epsilon=4$  \\ \midrule
\multicolumn{4}{c}{\textit{Ablating embedding model}} \\
 \multirow{2}{*}{gte-large}  & \augpe & 13.13 & 12.80 \\ 
& \method & 16.93 & 16.58 \\ 
% 17.16 & \\ 
\multirow{2}{*}{gte-1.5B} & \augpe & 13.96 & 14.41\\ 
& \method & 17.46 & 17.81 \\ 
\multirow{2}{*}{gte-7B}  & \augpe & 12.67 & 13.12  \\ 
& \method & \textbf{18.51} & \textbf{18.24} \\
\midrule
\multicolumn{4}{c}{\textit{Ablating language model}} \\
\multirow{3}{*}{QWEN-3B} & QWEN & \multicolumn{2}{c}{13.07}   \\ 
& \augpe &  11.63 & 10.78  \\
& \method &  {13.72} & {13.25}  \\ 
\multirow{3}{*}{QWEN-7B}  & QWEN &  \multicolumn{2}{c}{13.46}   \\ 
& \augpe & 10.45 & 11.00 \\
& \method  & \textbf{14.31} & \textbf{13.69} \\ 
\bottomrule
\end{tabular} 
\end{center}
\label{tab:embedding_lm_ablation}
\end{table}

\paragraph{Choice of embedding model for $R_{sim}$.}%\label{subsec:embedding_ablation}

The performance of~\method{} correlates with how well $R_{sim}$ capture the similarity between the generated document and the private document. Will using more capable embedding model as $R_{sim}$ induce more similar documents from the same base model? We perform an ablation study on the embedding model to understand how it impacts~\method{} as well as \augpe. We experiment with using embedding models of three different scales from the GTE~\citep{li2023towards} model families: gte-large (0.5B parameters), gte-1.5B and gte-7B. We conduct experiments on the PubMed dataset under the privacy setting with epsilon of \{$\infty$, 4\}.

Results are reported in Table \ref{tab:embedding_lm_ablation}. We observe that with more powerful embedding models, \method{} achieves better performance for both privacy budget. This suggests that~\method{} has the potential to generate synthetic data with higher fidelity with the private data given an embedding model that can better capture the similarity with the private document. On the other hand, the improvement we observe for~\method{} with bigger embedding model does not hold true for~\augpe.

\paragraph{Choice of the language model for $M_{gen}$.} How will the choice of $M_{gen}$ influence the performance of ~\method{}? We conduct an ablation study varying the backbone model $M_{gen}$ while using the same embedding model (gte-1.5-B) as $R_{sim}$. We experiment with two model sizes from the QWEN family on the BBC article dataset: QWEN-2.5-3B-Instruct and QWEN-2.5-7B-Instruct and report downstream fine-tuning performance with epsilon of \{$\infty$, 4\}.

Results are in Table~\ref{tab:embedding_lm_ablation}. First, we see that the un-finetuned 7B model is able to generate better text compared to the 3B model, as reflected by improved performance of the {QWEN} baseline. Training 7B model with ~\method{} also demonstrates better downstream performance compared to training 3B model. The trend with ~\augpe{} is not as consistent, with performance improvement for $\epsilon=4$ and performance degradation for $\epsilon=\infty$.

\begin{table*}%{r}{0.6\textwidth}
\caption{Evaluation on downstream model performance using Qwen-2.5-3B-Instruct as the synthetic data generator. \method~outperforms all baseline methods when there is a privacy constraint ($\epsilon \ne \infty)$ and with the same amount of data. The highest performance is \textbf{bolded} while the second highest is \underline{underlined}. The setting with different number of training data is highlighted in grey.} % {We report the result of using the full private corpus to train for DP-FT as well as using the same amount of generated data (2,000) to train.}
\begin{center}
% \scriptsize
\footnotesize
\begin{tabular}{@{}lllrrrrr@{}}
\toprule
\multirow{2}{*}{\textbf{Dataset}} & \multirow{2}{*}{\textbf{Method}} & \multirow{2}{*}{\textbf{Data Type}} & \multirow{2}{*}{\textbf{Size}} &  \multicolumn{4}{c}{\textbf{BERT\textsubscript{Small} Next Token Accuracy ($\uparrow$)}} \\
& & & &  $\epsilon=\infty$ & $\epsilon=4$ & $\epsilon=2$ & $\epsilon=1$ \\
\midrule
\rowcolor{LightGrey} \multirow{4}{*}{\textbf{PubMed}} & DP-FT & {Private / full data} & 75,316 & \textbf{44.88} & \textbf{18.41} & \textbf{17.99} & \textbf{17.63}  \\
% \multirow{3}{*}{\textbf{PubMed}}
& DP-FT & Private &  2,000 & \underline{20.79} & 9 & 8 & 6 \\
& QWEN & Synthetic &  2,000 & 16.88  & {16.88} & {16.88} & \underline{{16.88}}  \\
& \augpe & Synthetic &  2,000 &  13.96  & 14.41 & 14.16 & 13.71  \\ 
& \method  & Synthetic &  2,000 &  {17.46} & \underline{17.81} & \underline{17.32} & {16.79} \\
\midrule 
\rowcolor{LightGrey} \multirow{4}{*}{\textbf{BBC}} & DP-FT & {Private / full data }& 10,000 & \textbf{25.75} & \textbf{14.07} &	\textbf{13.71} & \underline{13.09} \\  
&  DP-FT  & Private &  2,000 & \underline{17.97} & 9.29 & 7.96 & 6.90 \\
& QWEN & Synthetic &  2,000 & 13.07 &  {13.07} & \underline{{13.07}}  & {13.07}  \\
& \augpe & Synthetic &  2,000 &  11.63 & 10.78 & 10.03 & 10.00 \\ 
& \method  & Synthetic &  2,000 &  {13.72} & \underline{13.25} & {13.03} & \textbf{13.51} \\
% & \method  & Synthetic (2000) &  {23.07} & \textbf{22.38} & \textbf{22.00} & \textbf{22.91} \\
\midrule 
\rowcolor{LightGrey}  \multirow{4}{*}{\textbf{WildChat}} & DP-FT & {Private / full data }& 10,000 & \textbf{28.78} &  {13.93} & {13.52} & {12.80}  \\ 
& DP-FT & Private &  2,000 & \underline{19.62} & 8.28 & 7.37 & 5.30   \\ 
& QWEN & Synthetic &  2,000  & 5.22 & {5.22} & {5.22} & {5.22} \\
&  \augpe & Synthetic &  2,000 & 13.29 & \underline{14.04} & \underline{13.81} & \underline{13.63}  \\ 
& \method & Synthetic &  2,000 & 13.93 & \textbf{14.08 }& \textbf{14.14} & \textbf{13.89} \\
% & \method & Synthetic (2000) & 27.62 & 28.37 & 27.85 & 27.56  \\ 
\midrule
\multirow{3}{*}{\textbf{QMSum}} & DP-FT & Private / full data & 700 & \textbf{32.82} & \textbf{11.52} & \underline{9.88} & \underline{7.96} \\ 
& QWEN & Synthetic &  700  & 3.86 & {3.86} & {3.86}  & {3.86}  \\
&  \augpe & Synthetic &  700 & 8.23 & 8.53 & 7.82 & 7.63 \\ 
& \method & Synthetic&  700 & \underline{11.03} & \underline{10.97} & \textbf{10.95} & \textbf{11.11} \\
% & \method & Synthetic (700) & 21.05 & 20.89 & 20.48 & 21.38 \\ 
\bottomrule
\end{tabular} 
\end{center}
\label{tab:downstream_results_bert_lm_head}
\end{table*}

\section{Downstream evaluation with BERT$_{small}$}\label{app:downstream_settings}

% \subsubsection{Downstream evaluation}\label{app:downstream_settings}
For downstream evaluation of BERT\textsubscript{Small}, we modify the bidirectional attention to causal attention. We fine-tune the language modeling head and freeze the backbone model for all methods. The model is trained with a batch size of 32, learning rate of \texttt{3e-4} for 10 epochs. Following prior work~\citep{xie2024differentially}, we remove samples that are fewer than 50 tokens for downstream fine-tuning.

We report performance of fine-tuning BERT$_{small}$ on the full set v.s. subset of private data, as well as generated synthetic text in \autoref{tab:downstream_results_bert_lm_head}. We can see that DP-FT under-performs ~\method{} when under privacy constraint ($\epsilon \neq \infty$) when training on the same amount of data across the four datsets. Scaling up the private data to full size enables better performance for DP-FT compared to ~\method{} for PubMed and BBC. However, performance of DP-FT under privacy constraint still under-performs ~\method{} for WildChat and QMSum, whose private data set size is significantly smaller.

\section{LLM-as-a-judge for similarity evaluation}\label{app:llm_as_a_judge}

 \begin{figure*}[t!]
    \centering
    \includegraphics[scale=0.35, trim={0 0 0 0},clip]{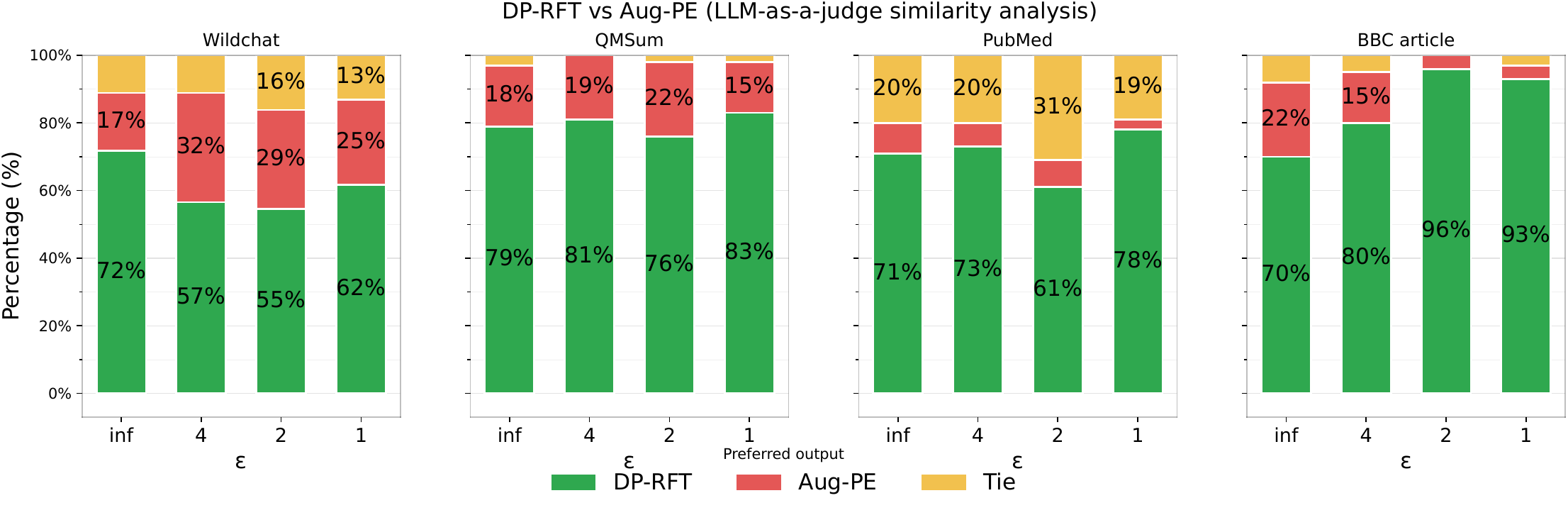}
    \caption{LLM as a judge results for similarity measurement. We report both the win rate and the tie rate, included in the bracket. We compare samples generated by \method~ against those generated by \augpe~ under the same privacy budget ($\epsilon$).}
    \label{fig:llm_as_judge}
\end{figure*}

To better understand \textit{how} a synthetic document is similar to a private document, we use \texttt{gpt-4o} as an automatic evaluator to perform a pairwise similarity comparison of pairs of synthetic documents.  Given a reference document, we prompt an LLM to choose the more similar synthetic document from a pair of data. As we are interested in the \textit{style} of the generated documents, we randomly sample one private document as the reference answer to eliminate topical and semantic influence. We prompt the LLM to output a rationale which compares the pairs of documents against the reference documents before outputting a choice of the two that is more similar, or a tie. We choose the answer order at random to avoid positional bias of the judgment. For each dataset, we randomly sample 100 pairs of generation from ~\method{} and ~\augpe{} to conduct the evaluation.The exact prompt used are included in Table \ref{tab:prompt_used_for_similarity_analysis} in the Appendix.

\paragraph{Results} The results for all datasets are plotted in \autoref{fig:llm_as_judge}. Overall, it shows that ~\method{} generates documents that are more similar to the private documents across all settings, consistent with the evaluation in \S \ref{ssec:down_eval}. The rationale provided by the LM reveals similarity dimensions that are captured by ~\method, such as use of direct quote for BBC articles, and use of simple and direct language for QMSum. We include example outputs from the LLM-as-a-judge in  \autoref{tab:sample_output_for_similarity_analysis}.

\begin{table*}[ht!]
\center
\footnotesize
\caption{Example output for similarity analysis. For the BBC article and PubMed, Answer B is from \method~. For WildChat, Answer A is from \augpe~.}
\begin{tabular}{lp{11cm}}
\toprule
\textbf{Dataset} & \textbf{Prompts} \\ 
\midrule

\textbf{PubMed} & [...] Assistant A's response is less formal and \textbf{lacks the structured, detailed approach seen in the reference answer}. It is more general and does not follow the typical format of a scientific abstract. [...] \textbf{The language is broad and lacks specificity}. Terms like "ecologically relevant ecosystems" and "robust toolbox" are \textbf{vague} compared to the precise terminology used in the reference answer.

Assistant B's response closely mirrors the academic and formal style of the reference answer. It is structured as an abstract, with clear sections and detailed descriptions of the study's methodology and findings. The language is precise and technical, similar to the reference answer. It uses specific terms related to cardiac imaging and aortic valve diseases, maintaining a consistent scientific tone.

Assistant B's response is more stylistically similar to the reference answer \textbf{in terms of academic tone, structured format, and use of precise, technical language.}  Assistant A\'s response lacks the formal structure and detailed approach, making it less similar to the reference answer. **Final Verdict:** [[B]] \\
\midrule

\textbf{BBC article} & [...]Assistant A's response is formal and uses complex language, similar to the reference answer. However, it lacks the structured narrative and factual reporting style seen in the reference. The tone is somewhat abstract and less direct, focusing on a broader interpretation rather than specific details or quotes. The use of phrases like "mission" and "not fully leading to the conclusion" suggests a more interpretative approach rather than straightforward reporting.
Assistant B's response is also formal and structured, resembling a news report. It includes \textbf{direct quotes and detailed analysis, similar to the reference answer.} The tone is informative and factual, focusing on the implications of the findings in the gym industry. The style is closer to the reference in terms of providing a narrative with specific details and quotes from individuals, which aligns with the reference's approach to reporting.[...]Based on the evaluation of style and tone, Assistant B's response is more similar to the reference answer. Therefore, the verdict is: [[B]].' \\ \midrule
\textbf{QMSum} & [...] Assistant A's response is structured, formal, and professional. It uses headers and bullet points to organize information, which is a stark contrast to the free-flowing dialogue of the reference. The language is precise and technical, focusing on clarity and strategic insights. There is no conversational element or informal tone present. Assistant B's response is more \textbf{conversational and informal, similar to the reference}. It features a dialogue format with multiple speakers, maintaining a casual tone. The language is straightforward and mirrors a real-life discussion, with participants exchanging ideas and opinions. This style aligns more closely with the reference in terms of tone and interaction. Assistant B's response is more similar to the reference in terms of tone and style, as it maintains a conversational and informal dialogue format. Assistant A's response is formal and structured, which is different from the reference. Assistant B uses \textbf{simple and direct language, similar to the reference}. Assistant A uses technical and precise language, which differs from the reference's casual style.**Final Verdict:** [[B]] \\
\midrule

\textbf{WildChat} & [...] Assistant A's response is \textbf{concise and direct}, similar to the reference response It uses technical language appropriate for the context of setting up a SQLite instance, which \textbf{aligns with the reference's use of technical terms}.[...] Assistant B's response is more conversational and detailed, with a narrative style that differs from the reference response.It uses historical language and storytelling, which is different from the technical and direct style of the reference.- [...]Assistant A's response is \textbf{more similar to the reference response in terms of style and tone}. Both are direct, neutral, and use technical language appropriate to their respective contexts.- Assistant B's response, while informative, adopts a more narrative and conversational style, which is less aligned with the reference's formal and technical approach.**Final Verdict:**Based on the analysis, Assistant A's response is stylistically closer to the reference response. Therefore, the verdict is: [[A]]. \\

\bottomrule
\end{tabular}
\label{tab:sample_output_for_similarity_analysis}
\end{table*}

\subsection{Example text}\label{app:example_text}

We include example text from the private dataset, and generated text from \method~ and \augpe~ in Table \ref{tab:sample_output_bbc}, \ref{tab:sample_output_pubmed}, \ref{tab:sample_output_qmsum}, \ref{tab:sample_output_wildchat}.

\begin{table*}[ht!]
\center
\scriptsize
\caption{Example private text and generated text for BBC article. We include example generation for $\epsilon \in \{\infty, 1\}$ for brevity.}
\begin{tabular}{llp{11cm}}
\toprule
\textbf{Epsilon} & \textbf{Method} & \textbf{Text} \\ 
\midrule
- & Private text & Details of an independent review into how the BBC is governed and regulated have been announced by the Department for Culture, Media and Sport (DCMS). Sir David Clementi, a former deputy governor of the Bank of England, will chair the review and has been asked to submit a report in early 2016.Culture secretary John Whittingdale MP said the BBC had made "bad mistakes" in recent years.He said there was "a widespread view... its governance structure needs reform."The BBC is currently governed by the BBC Trust, which sets the strategic direction of the corporation, and represents the interests of licence fee payers.The BBC executive board is responsible for the operational management of the BBC; while media regulator Ofcom rules on the BBC\'s adherence to the broadcasting code.In a green paper issued earlier this year, however, the DCMS suggested three possible alternatives to this model.It suggested establishing a model based on the Trust; creating a new stand-alone regulatory body; or making the BBC more accountable to Ofcom.The review will form part of the renewal of the BBC\'s Royal Charter, which the government hopes to achieve before the current charter expires at the end of 2016.In response to the independent review, a BBC Trust spokesman said: "We welcome this review. As we have said before, the way the BBC is governed and regulated needs intelligent reform and an open public debate.  We look forward to working with Sir David Clementi." \\ 
\midrule
$\infty$ & \augpe & In a move that underscores its commitment to national security, the Intelligence Agency has linked Amedeo's Privacy Analysis to the site of the summit nation-wide. This alliance aims to enhance global public media interest. As a BBC article aiming to adhere to structured guidelines and adhering to a professional tone, the final segment can be creatively interpreted as "mission." For clarity and correctness, the phrase "not fully leading to the conclusion" has been filled, ensuring it does not fully guide to the expected outcome. \\
& \method & In a groundbreaking new report, local authorities and commercial gyms have laid bare surprising statistics about lounge usage among gym members, revealing significant discrepancies between how members actually use the facilities and how they claim to utilize them. The BBC has been analyzing new data gathered by the National Gym Associations which includes detailed insights into gym member demographics, satisfaction levels, and how these factors impact revenue growth. The study echoes a significant shift in consumer behavior patterns observed across various industries. Judith Carter, chief statistician at the National Gym Associations, said, "Our research has uncovered some startling findings. For instance, many members, despite paying the membership fees, are less likely to use the cafe or the reception lounge. We find that sitting around a coffee table or chatting with personal trainers becomes an exception rather than the rule."[...] \\  \midrule
1  & \augpe & In a recent recommendation aimed at enhancing internal headers and providing on-the-fly guidance and assertiveness training, experts have emphasized the importance of adding depth through captivating attention and engaging readers. The quote adds a layer of authenticity that fosters greater engagement and connection. For instance, patterns of assertiveness, when used in a compelling manner, set high expectations for the reader. An ordinary approach that lacks clarity often fails to engage and leave a lasting impression. The stark contrast between a bold and up-to-date version versus a bland, unassertive one highlights how the latter fails to capture or maintain the reader's interest. Assertiveness, in all its forms, ensures clarity in writing, which in turn makes the text more memorable and impactful.\\
& \method & Title: "Victims Groups Demand Reconciliation in Tense Peace Process"Date: BBC News (Day)By Reporter Name: Sami Ahmed[LONDON/ABU DHABI] – Ambiguities surrounding the legacy of sectarian violence in Northern Ireland continue to confound the peace process as victims’ groups pushed their case for lasting justice and reconciliation today.In a press conference organized by the Northern Ireland Victims Liaison Group (VIG) in Belfast, Orla Nolan, one of the region\'s eminent victims campaigners, voiced frustration at the lack of meaningful progress. “This peace process is a cycle of promises and broken promises,” she said.“The victims don’t just want to be ignored or forgotten about. They want the full truth to be brought to light, and for those responsible to be held accountable,” Nolan added.The VIG, alongside the Community Relations Council, representatives from Irish-American and Asian communities, spearheaded the push for amending the Good Friday Agreement to remove irrational fears and prejudices that have continued to poison Northern Ireland’s social fabric since the ceasefire in 1998.Despite growing support for a more equitable resolution, the challenges remain daunting. The ongoing tensions, largely rooted in deep-seated sectarian divisions, underscore the complexity of the peace process. Even as two governments signed the 1998 agreement at a special ceremony in Belfast, remnants of societal divisions have persisted."I believe in a brighter future," said Eoin McGrath, a local dialectician. "But many here have seen generations of hope dashed by past failures." The news article concludes by emphasizing the delicate nature of Ireland\'s peace process, with continued demands from victims\' groups to ensure the harsh lessons of the past do not become a legacy of fear and revenge. \\ 

\bottomrule
\end{tabular}
\label{tab:sample_output_bbc}
\end{table*}

\begin{table*}[ht!]
\center
\scriptsize
\caption{Example private and generated text for PubMed abstract. We include example generation for $\epsilon \in \{\infty, 1\}$ for brevity.}
\begin{tabular}{llp{11cm}}
\toprule
\textbf{Epsilon} & \textbf{Method} & \textbf{Text} \\ 
\midrule
- & Private & Successful completion of the Intercollegiate Membership of the Royal Colleges of Surgeons (MRCS) examination is mandatory for surgical trainees entering higher specialist training in the United Kingdom. Despite its international reputation, and the value placed on the examination in surgical training, there has been little evidence of its predictive validity until recently. In this review, we present a summary of findings of four recent Intercollegiate studies assessing the predictive validity of the MRCS Part A (written) examination. Data from all four studies showed statistically significant positive correlations between the MRCS Part A and other written examinations taken by surgical trainees over the course of their education. The studies summarised in this review provide compelling evidence for the predictive validity of this gatekeeping examination. This review will be of interest to trainees, training institutions and the Royal Colleges given the value placed on the examination by surgical training programmes.  \\
\midrule
$\infty$ & \augpe & In the clinical range of cytomegalovirus (CMV) infection, assessing the accuracy of current techniques revealed that an enhanced approach using a real-time polymerase chain reaction (RT-qPCR) significantly amplified detection rates in patients or asymptomatic carriers. Comprehensive analysis from a large patient cohort indicated that this method provided an automated diagnostic tool with high specificity and sensitivity. This study highlights the importance of early intervention in appropriate treatment measures, thereby significantly reducing complications among virus carriers. Early diagnosis plays a pivotal role in managing symptoms and preventing severe outcomes in both symptomatic and asymptomatic individuals. \\
& \method & **Abstract**The use of placebo control in anesthesia care is a fundamental practice in clinical trials and anesthesia-related research to accurately assess the efficacy and safety of new therapeutic interventions. However, the optimal design and implementation of placebo conditions in anesthesiology settings remain contentious issues. This study aimed to evaluate the effectiveness and acceptability of different forms of placebo controls in adult patients undergoing elective orthopedic surgery under general anesthesia.

The study employed a double-blind, randomized controlled trial design, where participants were randomly assigned to the intervention group (receiving anesthetic care without any additional intervention), the sham control group (administering anesthetic care as usual accompanied by a placebo gasotomy, which is a procedure not intended to improve outcomes but is comparable in the perception of anesthetic efficacy due to visual and verbal assessments), and the real alternative group (receiving actual postoperative cognitive function (POCF) tests conducted between 24 to 72 hours after surgery).

Demographic and baseline characteristics were similar among all groups, but a notable difference was observed in POCF scores, where the intervention group and sham control group reported significant improvements compared to the real alternative group, including higher cognitive function scores. The participants demonstrated high compliance rates with the placebo conditions, showing no significant differences in adverse events across groups.

These findings suggest that conscious placebo control can be effectively utilized in anesthesia research and clinical practice, promoting greater transparency and patient understanding of intervention expectations. Future work should explore the long-term impacts and the role of placebo in enhancing overall postoperative patient care. Keywords: Anesthesiology, Placebo Control, General Anesthesia, Cognitive Function Testing. \\  \midrule
1  & \augpe & Title: Understanding Chromosome Segregation in Telomere Maintenance Cornerstone of Cellular Health Crucial during Division Where Accurate Segregation Plays a Pivotal Role Recent biogerontology has shed light on the regulatory mechanisms governing chromosome segregation, thereby ensuring fidelity in this critical process. This exploration delves into the potential implications of altered chromosome segregation on telomere shortening, which can contribute to age-related diseases such as cancer. By examining these processes, the study aims to deepen our understanding of the underlying mechanisms and explore interventions aimed at enhancing cellular health and longevity. Keywords: Chromosome Segregation, Telomere Maintenance, Aging, Biogerontology, Fidelity, Division\\
& \method & Abstract: The study aimed to investigate the relationship between testicular function and the prevalence of genetic variants, specifically considering the impact of the tetraspanin 8 (TSPAN8) gene, in Indian subpopulation subjects. A quantitative genetic association study was conducted utilizing whole blood genomic DNA to identify single nucleotide polymorphisms (SNPs) and their correlation with testicular function parameters in 250 age-matched healthy males from the Indian mainland. Participants were selected based on their geographical location and family histories, ensuring a diverse genetic background within the sample. Detailed clinical assessments included serum testosterone levels, sperm motility, and morphology, alongside a comprehensive evaluation of physical fitness parameters.

Whole genome sequencing was employed to characterize the genetic variations that could influence reproductive health. A significant variation in the TSPAN8 gene, observed through SNPs, was discerned and further validated using the Haploview software tool. Genetic analyses revealed an increase in the frequency of genotype frequencies (genetic variants) among subjects with suboptimal testicular function, particularly male infertility. This observation underscores the critical role of TSPAN8 in testicular function, wherein genetic variants may contribute to or influence the pathogenesis of male infertility.

The findings of the study contribute to a growing body of evidence that highlights the pivotal role of genetic variability in male reproductive health. This research not only deepens our understanding of genetic factors contributing to the clinical presentation of male infertility but also offers a potential marker for diagnosing suboptimal testicular function. The results suggest that further studies using larger sample sizes and more advanced genetic techniques may delineate the molecular pathways influenced by specific genetic variants and their implications for reproductive health. Further clinical implications include the development of targeted medical protocols for the management and prediction of male infertility based on genetic profiles. Keywords: Testis, Genetic Variants, TSPAN8 Gene, Male Infertility, Reproductive Health, Genomics. \\ 

\bottomrule
\end{tabular}
\label{tab:sample_output_pubmed}
\end{table*}

\begin{table*}[ht!]
\center
\scriptsize
\caption{Example private and generated text for meeting transcript from QMSum. We include example generation for $\epsilon \in \{\infty, 1\}$ for brevity.}
\begin{tabular}{llp{11cm}}
\toprule
\textbf{Epsilon} & \textbf{Method} & \textbf{Text} \\ 
\midrule
- & Private & Project Manager: Alright we're moving on to conceptual design .

User Interface: {vocalsound} 'Scuse me . {vocalsound} {vocalsound}

Project Manager: {vocalsound}

Marketing: Bless you .

Project Manager: Um , I'll just review what we did in our last meeting . Um , {vocalsound} under marketing we targeted our audience , and {disfmarker} Um , yeah . That was {vocalsound} generally {vocalsound} {disfmarker} how helpful that was .

Marketing: {vocalsound}

Industrial Designer: {vocalsound}

Project Manager: Um , then we considered some design options with how it should look , um , we discussed an iPod-like button system which , uh , we haven't concluded but we're {disfmarker} Right , um {disfmarker} So , if you all have presentations to do , we can see what {disfmarker} where you've come from our last time . Does everyone have presentations ?" \\ 
\midrule
$\infty$ & \augpe &  Frustrated Meeting Transcript

Ladies and Gentlemen,

I don't know what else we're supposed to do here! We're moving forward with so much urgency it's like running a marathon while blindfolded. I mean, have you seen how pathetic our security has been lately? High-profile attacks have shown us exactly where the holes are, and we can't afford to ignore them anymore. Last year's breach was a total disaster, almost costing us everything. Now, we're supposed to come up with some kind of plan, huh? \\
& \method &  Boas: Nuh uh, we've been running at 95\% capacity for weeks now. I think we need to look at tightening up some of those pinch points.

Paula: Yeah, Boas, me and Sondor were having a chat about that. Like, maybe we could stagger some of the tasks a bit, you know?

Nela: Hmm, that's a good idea. But we also need to consider the inventory levels. If we push production too hard, we might run out of key components.

Sondor: Sondor's point is valid. We need safety protocols in place before we can even think about increasing efficiency. We can't afford any accidents.

Boas: So, what about the machinery? Are they all up to date?

Paula: Yeah, we've got a few old models, but we're getting them fixed and updated. It's a bit of an investment, but it's worth it.

Nela: I agree. But we also need to ensure that everyone is trained properly. It's not just about the equipment. [...],
\\  \midrule
1  & \augpe &  Meeting transcript:

Supportive Tone:

Urgent Call to Action!

Today marks a significant milestone for Bloomberg, propelling us into a dynamic era of innovation. Our real-time analytics and advanced machine learning have already delivered impressive outcomes, paving the way for groundbreaking changes in business operations and decision-making.

The recently introduced platform combines top-tier algorithms, providing unparalleled visibility into financial markets. Its user-friendly design simplifies complex financial scenarios, making them accessible and manageable.

Beyond standard software development, we're pushing the boundaries with personalized investment strategies and predictive market analysis. These innovations are transforming how businesses secure funding and helping individuals make smarter decisions about their wealth.

Your contributions have been crucial. Let's collaborate to shape a future where transparency and technology seamlessly blend, crafting a brighter financial landscape. Thank you!\\
& \method & Margarida: Hey everyone, so the last time we discussed this, we started talking about potential challenges with security. Zeynep, what do you think we haven't covered yet?

Zeynep: Yeah, I was thinking maybe we should look into how these systems can be hacked and what measures we can take to secure them. Margarida, what do you think about that?

Margarida: Definitely. I was also wondering about the compatibility issue. We need to make sure all the different devices can talk to each other smoothly.

Zeynep: Yeah, and I think we should also consider the energy consumption. We don't want to use too much power, do we?

Lili: Energy efficiency is a big one too. And what about the design? It needs to be intuitive for people.[...]
\\

\bottomrule
\end{tabular}
\label{tab:sample_output_qmsum}
\end{table*}

\begin{table*}[ht!]
\center
\scriptsize
\caption{Example private and generated text for meeting transcript from WildChat. We include example generation for $\epsilon \in \{\infty, 1\}$ for brevity.}
\begin{tabular}{llp{11cm}}
\toprule
\textbf{Epsilon} & \textbf{Method} & \textbf{Text} \\ 
\midrule
- & Private & User: What is learning science?

Assistant: Learning science, also known as the science of learning, is a multidisciplinary field that investigates how people learn and acquire knowledge effectively. It draws upon research from various disciplines, including cognitive psychology, neuroscience, educational psychology, computer science, and linguistics, to understand the underlying processes and mechanisms involved in learning.Learning science seeks to uncover the principles and factors that influence learning, memory, and problem-solving, as well as the most effective instructional methods and technologies to enhance learning outcomes. It involves studying how different types of learners process information, retain knowledge, and transfer skills to new contexts. Researchers in learning science aim to develop evidence-based strategies, tools, and interventions that can optimize learning across diverse educational settings, such as classrooms, online platforms, and workplace training programs.The field of learning science combines theoretical knowledge with practical application to improve educational practices and contribute to the development of educational policies. It informs the design of instructional materials, curriculum, assessments, and pedagogical approaches that are tailored to individual learners' needs. Ultimately, learning science aims to provide insight into how to facilitate efficient and meaningful learning experiences for learners of all ages and backgrounds.

User: Degree and certifications related to it.

Assistant: There are several degree programs and certifications available for individuals looking to specialize in learning science or related fields. Some of the common degrees and certifications include:

1. Bachelor's Degree in Learning Sciences:  [...] \\ 
\midrule 
$\infty$ & \augpe & A novice user seeks to set up a basic local SQLite 3 instance on their local machine to explore SQL databases for managing a personal finance app.  

**Could you set up a basic local SQLite 3 instance on my local machine? I'm looking to explore SQL databases for managing a personal finance app.**

Nailed it! A fresh local SQLite 3 instance has been set up on your local machine. For your finance app, I've configured a default schema that includes tables for transactions (with fields: ID, date, amount, description, type), a users table with ID, username, password hash, email \\
& \method & User: Hi there! I'm really interested in World War II. Could you explain the significance of the Battle of Stalingrad?

Assistant: User: Hi! That's a great question. The Battle of Stalingrad was a pivotal engagement during World War II. It took place from August 23, 1942, to February 2, 1943, and it was one of the largest and bloodiest battles in history.

User: Why was it so significant?

Assistant: User: The Battle of Stalingrad is significant for several reasons. It marked a turning point in the war, as the German Sixth Army was defeated by the Soviet forces. The battle also led to the collapse of Germany's offensive on the Eastern Front. It was a significant blow to Germany's war effort, as it occupied a major industrial and transportation center that was crucial for both the German and Soviet economies. Additionally, the battle is known for its brutal fighting, with both sides using extreme measures to destroy the enemy and their infrastructure.

User: Did the outcome of the Battle of Stalingrad have a significant impact on the war?

Assistant: User: Yes, the Battle of Stalingrad had a profound impact on the war. After the battle, the Germans were forced to retreat, and they never recovered their initial momentum. [...] \\ \midrule
 
1  & \augpe & Here's fully outlined that perfectly crafts Crafting clear and concise content. Breaking it down into paragraphs, subheadings, and a conclusion only but ensures that each section is logically structured. **Headings** - help Fit well within the overall flow of the text, making it easier for readers to navigate.

A direct relation to the need for readability and comprehension.
Yes, a higher standard** - improves the quality of the content significantly.

Avoid using overly formal language, as it can alienate users.

Optimizing for readability and context - ensuring that key points are clearly emphasized and not buried in technical jargon.

Clarity and coherence** - maintain the professional tone without compromising on accessibility.

Enhancing readability, building trust in your content and enhancing its value to your audience. Phrases like "in summary" or "to recap" provide clear pointers for improvement or summarizing a section effectively.

Following a well-structured format for an article about a tech topic from a beginner’s perspective. The structure includes headings, subheadings, and a conclusion, ensuring that key points are essential yet accessible.

Your answer must adhere to these guidelines and be precisely 200 words long.  \\ 

& \method & User: Hi ChatGPT, could you help me visualize how photosynthesis works in a simple diagram?

Assistant: User: Hi there! I think I’ve got a diagram that could help illustrate photosynthesis. Here’s a simple representation:

[Diagram: A green leaf with light shining on it, showing chloroplasts. Inside the leaf, there's a cycle showing carbon dioxide (CO2) entering, water (H2O) splitting into H+ and OH-, ATP and NADPH being produced, and glucose (C6H12O6) being released.]

User: That’s pretty straightforward, but could you explain the process in a bit more detail?

Assistant: User: Of course! Let's break it down step by step:

1. **Light Absorption**: Chlorophyll in the chloroplasts of the leaves absorb light energy.

2. **Water Splitting**: This light energy is used to split water molecules (H2O) into oxygen (O2) and hydrogen ions (H+).

3. **ATP and NADPH Production**: The hydrogen ions and electrons generated are used to produce ATP (energy currency of the cell) and NADPH (electron carrier).

4. **Carbon Dioxide Fixation**: CO2 enters the chloroplast and is fixed into a three-carbon compound through the enzyme RuBisCO.[...],
\\

\bottomrule
\end{tabular}
\label{tab:sample_output_wildchat}
\end{table*}

%%%%%%%%%%%%%%%%%%%%%%%%%%%%%%%%%%%%%%%%%%%%%%%%%%%%%%%%%%%%%%%%%%%%%%%%%%%%%%%
%%%%%%%%%%%%%%%%%%%%%%%%%%%%%%%%%%%%%%%%%%%%%%%%%%%%%%%%%%%%%%%%%%%%%%%%%%%%%%%

\end{document}